\documentclass[runningheads]{llncs}
\usepackage{graphicx}

\usepackage{tikz}
\usetikzlibrary{calc}

\usepackage{comment}
\usepackage{amsmath,amssymb} %

\usepackage{booktabs}
\usepackage{epigraph}
\usepackage{multirow}
\usepackage{makecell}
\usepackage{epsfig}
\usepackage{xcolor}
\usepackage{xspace}
\usepackage{subcaption}
\usepackage{hyperref}

\usepackage[leftmargin=1em]{quoting}
\usepackage{mdframed}
\usepackage{enumitem}
\mdfdefinestyle{TaskFrame}{
    linecolor=gray,
    linewidth=0.25pt,
    innertopmargin=3pt,
    innerbottommargin=3pt,
    innerrightmargin=3pt,
    innerleftmargin=3pt,
    leftmargin = -2pt,
    rightmargin = -2pt,
    backgroundcolor= black!03}

\usepackage[capitalize]{cleveref}
\crefname{section}{Sec.}{Secs.}
\Crefname{section}{Section}{Sections}
\Crefname{table}{Table}{Tables}
\crefname{table}{Tab.}{Tabs.}

\usepackage[accsupp]{axessibility}  %

\definecolor{mycolor}{RGB}{229, 83, 0}

\newcommand\tildetext{{\raise.17ex\hbox{$\scriptstyle\mathtt{\sim}$}}}

\newcommand\sherlock{\textbf{\textsc{Sherlock}}\xspace}

\newcommand\confidence{\textsc{confidence score}\xspace}
\newcommand\observation{\textsc{observation pair}\xspace}
\newcommand\observations{\textsc{observation pairs}\xspace}

\newcommand\clue{clue\xspace}

\newcommand\clues{clues\xspace}
\newcommand\Clues{Clues\xspace}
\newcommand\inferences{inferences\xspace}
\newcommand\Inferences{Inferences\xspace}
\newcommand\inference{inference\xspace}
\newcommand\inferencewithboldedf{in\underline{\textbf{F}}erence\xspace}
\newcommand\Inference{Inference\xspace}

\newcommand\retrieve{retrieve\xspace}

\newcommand\retrieval{retrieval\xspace}
\newcommand\Retrieval{Retrieval\xspace}
\newcommand\Retrievallongtitle{Retrieval of Abductive Inferences\xspace}
\newcommand\Retrievallongsent{Retrieval of abductive inferences\xspace}

\newcommand\localize{localize\xspace}

\newcommand\localization{localization\xspace}
\newcommand\Localization{Localization\xspace}
\newcommand\Localizationlongtitle{Localization of Evidence\xspace}
\newcommand\Localizationlongsent{Localization of evidence\xspace}

\newcommand\compare{compare\xspace}

\newcommand\comparison{comparison\xspace}
\newcommand\Comparison{Comparison\xspace}
\newcommand\Comparisonlongtitle{Comparison of Plausibility\xspace}
\newcommand\Comparisonlongsent{Comparison of plausibility\xspace}

\newcommand\fig{Fig.\xspace}

\newcommand*{\img}[1]{%
    \raisebox{-.15\baselineskip}{%
        \includegraphics[
        height=\baselineskip,
        width=\baselineskip,
        keepaspectratio,
        ]{#1}%
    }%
}

\usepackage{orcidlink}

\begin{document}
\pagestyle{headings}
\mainmatter
\def\ECCVSubNumber{5263}  %

\title{\img{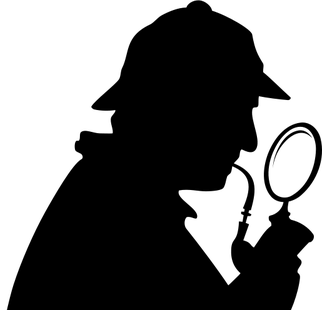} \emph{The Abduction of Sherlock Holmes:} \\ A Dataset for Visual Abductive Reasoning} %

\titlerunning{\img{img/sherlock_image.png} Sherlock: A Dataset for Visual Abductive Reasoning}
\author{Jack Hessel*\inst{1}\orcidlink{0000-0002-4012-8979} \and
Jena D. Hwang*\inst{1}\orcidlink{0000-0003-3801-294X}\index{Hwang, Jena D.} \and
Jae Sung Park\inst{2}\orcidlink{0000-0003-3678-6925}\index{Park, Jae Sung} \and
Rowan Zellers\orcidlink{0000-0003-1361-9868}\inst{2} \and
Chandra Bhagavatula\inst{1}\orcidlink{0000-0001-6264-0378} \and
Anna Rohrbach\inst{3}\orcidlink{0000-0003-1161-6006} \and
Kate Saenko\inst{4}\orcidlink{0000-0002-5704-7614} \and \\
Yejin Choi\inst{1,2}\orcidlink{0000-0003-3032-5378}
}
\authorrunning{J. Hessel et al.}
\institute{Allen Institute for AI  \email{\{jackh,jenah,chandrab\}@allenai.org} \and
Paul G. Allen School of Computer Science \& Engineering, University of Washington\\ 
\email{\{jspark96,rowanz,yejin\}@cs.washington.edu} \and
University of California, Berkeley \email{anna.rohrbach@berkeley.edu} \and
Boston University and MIT-IBM Watson AI  \email{saenko@bu.edu}}
\maketitle

\begin{abstract}
Humans have remarkable capacity to reason abductively and hypothesize
about what lies
beyond the literal content of an image. By identifying concrete visual
\textbf{\clues} scattered throughout a scene, we almost can't help but draw probable \textbf{\inferences} beyond the literal scene based on our everyday experience and knowledge about the world. For example,
if we see a ``20 mph'' sign alongside a road, we might assume the street sits in a residential area (rather than on a highway), even if no houses are pictured. 
Can machines perform similar visual reasoning?

We present \sherlock, an annotated corpus of 103K images for testing machine capacity for abductive reasoning \emph{beyond} literal image contents. We adopt a free-viewing paradigm: participants first observe and identify salient \clues within images (e.g., objects, actions) and then provide a plausible \inference 
about the scene, given the \clue. In total, we collect 363K (\clue, \inference) pairs, which form a first-of-its-kind 
abductive visual reasoning dataset.
Using our corpus, we test three complementary axes of abductive reasoning. We evaluate the capacity of models to: i) \emph{\retrieve} relevant \inferences from a large candidate corpus; ii) \emph{\localize} evidence for inferences via bounding boxes, and iii) \emph{\compare} plausible inferences to match human judgments on a newly-collected diagnostic corpus of 19K Likert-scale judgments. While we find that fine-tuning \texttt{CLIP-RN50x64} with a multitask objective outperforms strong baselines, significant headroom exists between model performance and human agreement. Data, models, and leaderboard available at \url{http://visualabduction.com/}.

\end{abstract}

\vspace{-4mm}
\epigraph{You know my method.\\It is founded upon the observation of trifles.}{\textit{``The Boscombe Valley Mystery'', by A. C. Doyle}}
\vspace{-4mm}

\section{Introduction}

\begin{figure}[t!]

\begin{center}
\includegraphics[width=1\linewidth]{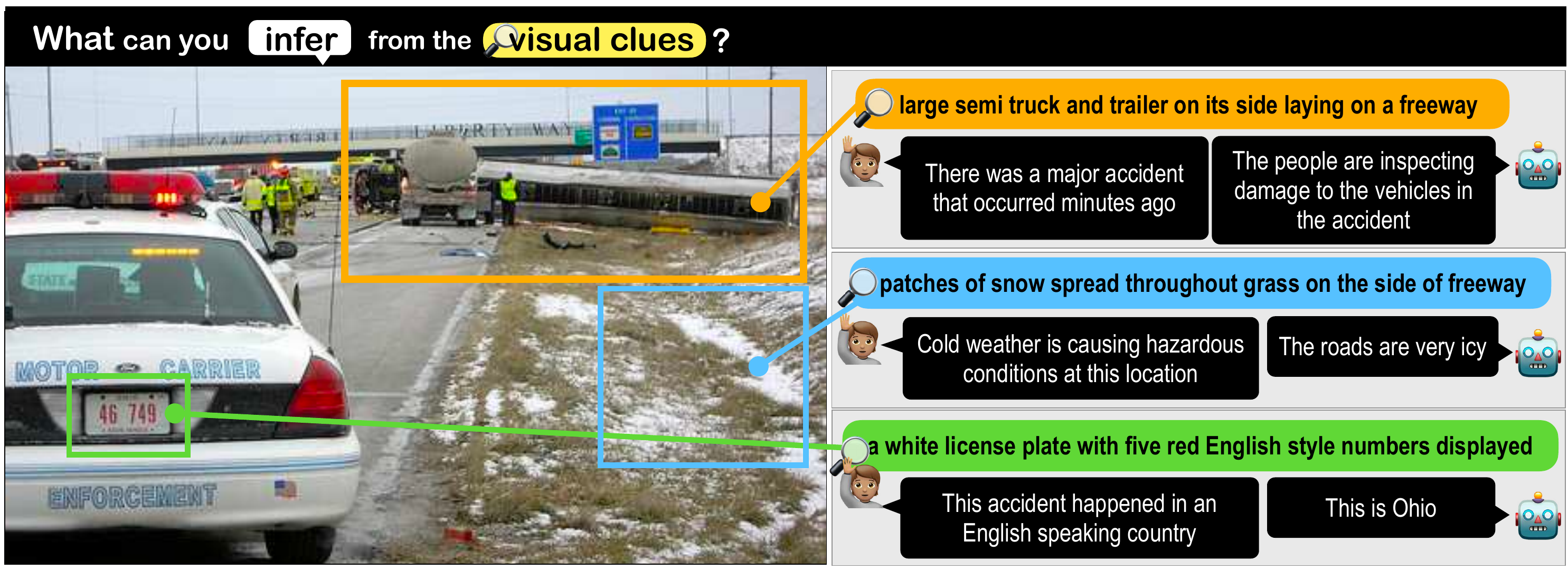}
\end{center}
\caption{%
We introduce \sherlock: a corpus of 363K commonsense inferences grounded in 103K images. Annotators 
highlight
localized \clues (color bubbles) and draw plausible abductive \inferences about them (speech bubbles). Our models are able to predict localized inferences (top predictions are shown), but we quantify a large gap between machine performance and human agreement.}

\label{fig:first_examples}
\end{figure}

The process of making the most plausible inference in the face of incomplete information is called \emph{abductive reasoning,} \cite{peirce1965pragmatism} personified by the iconic visual inferences of the fictional detective Sherlock Holmes.\footnote{While Holmes rarely makes mistakes, he frequently misidentifies his mostly abductive process of reasoning as ``deductive.'' \cite{niiniluoto1999defending,carson2009abduction}}
Upon viewing a scene, humans can quickly synthesize cues to arrive at abductive hypotheses that go beyond the what's captured in the frame. 
Concrete cues are diverse: people take into account the emotion and mood of the agents, speculate about the rationale for the presence/absence of objects, and zero-in on small, contextual details; all the while accounting for prior experiences and (potential mis)conceptions.\footnote{
The \emph{correctness} of abductive reasoning is certainly 
not guaranteed. Our goal is to study perception and reasoning \emph{without} endorsing specific inferences (see \S \ref{sec:sec_with_use_cases}).%
}
\fig~\ref{fig:first_examples} illustrates: 
snow may imply dangerous road conditions, 
an Ohio licence plate may suggest the location of the accident, and a blue sign may indicate this road is an interstate. 
Though not all details are equally important, certain \emph{salient} details shape
our abductive %
inferences about the scene as a whole
\cite{shank1998extraordinary}. This type of visual information is often left
unstated.

\begin{figure}[t]
\begin{minipage}{0.65\textwidth}
    \centering
        
    \resizebox{\linewidth}{!}{
        \includegraphics[width=1\textwidth]{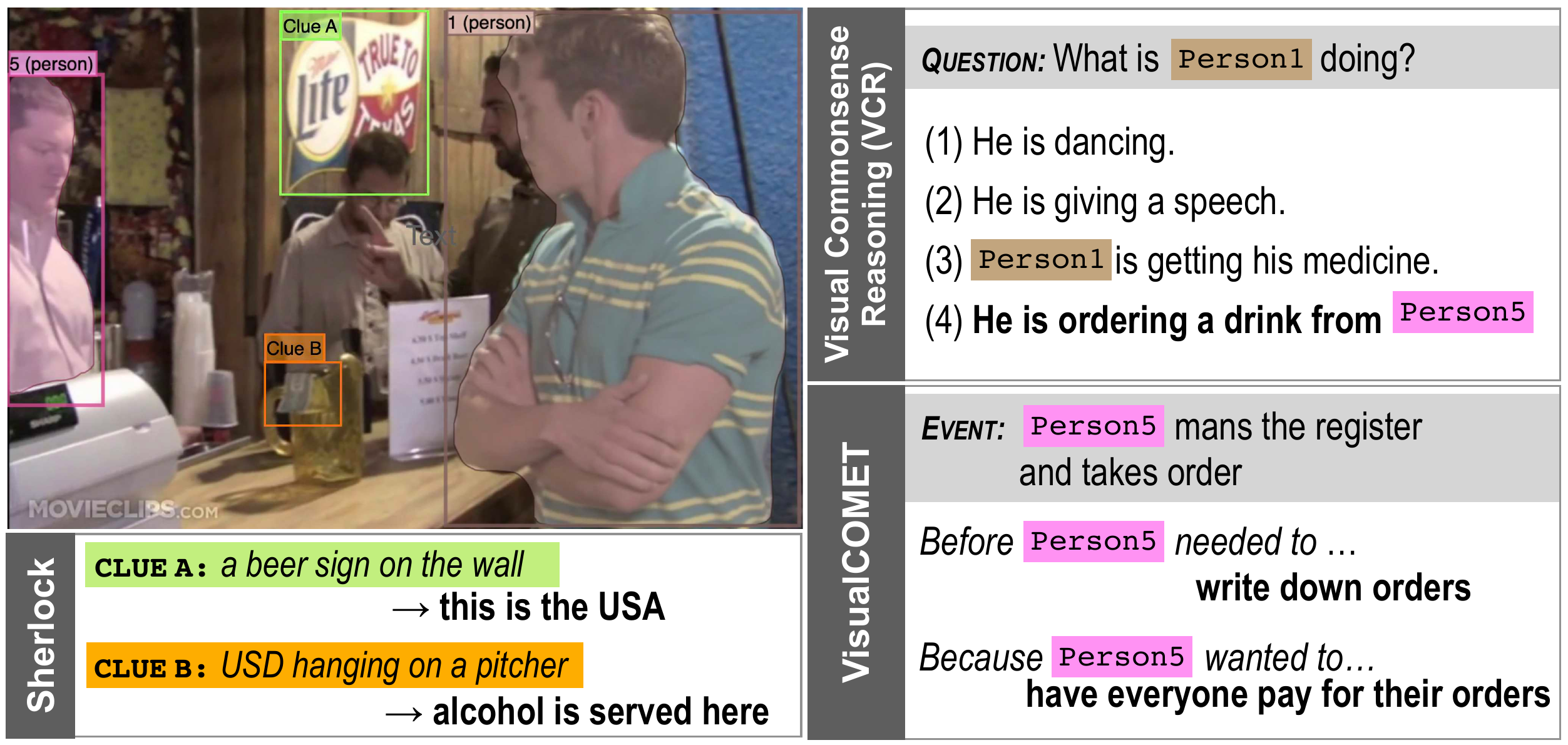}
    }
    \vspace{1mm}
\end{minipage}
\hfill
\begin{minipage}{0.33\textwidth}
    
    \caption{Side-by-side comparison of VCR~\cite{zellers2019recognition}, VisualCOMET~\cite{park2020visualcomet}, and \sherlock on a representative instance. \sherlock showcases a wider range of (non-human centric) situational contexts.
    }
    \label{tab:prior_work_comparison}    
\end{minipage}
\end{figure}

We introduce \sherlock, a new dataset of 363K commonsense inferences grounded in 103K images. %
\sherlock makes explicit typically-unstated cognitive processes: each image is annotated with at least 3 inferences which pair depicted details (called \clues) with commonsense conclusions that aim to go \emph{beyond} what is literally pictured (called \inferences). \sherlock is more diverse than many existing visual commonsense corpora like Visual Commonsense Reasoning \cite{zellers2019recognition} and VisualCOMET \cite{park2020visualcomet},\footnote{For instance, 94\% of visual references in \cite{zellers2019recognition} are about depicted actors, and \cite{park2020visualcomet} even requires KB entries to explicitly regard people; see \fig~\ref{tab:prior_work_comparison}.} due to its free-viewing data collection paradigm: we purposefully do not pre-specify the types of \clues/\inferences allowed, leaving it to humans to identify the most salient and informative elements and their implications.
Other forms of free-viewing like image captions may not be enough: a typical caption for \fig~\ref{fig:first_examples}
 may mention the accident and perhaps the snow, but
smaller yet important details needed to comprehend the larger scene (like the blue freeway sign or the Ohio plates) may not be mentioned explicitly \cite{berg2012understanding}.
Dense captioning corpora \cite{johnson2016densecap} attempts to overcome this problem by
highlighting
\emph{all} details, but it does so without
accounting for which details are salient (and why).

\begin{table}[t]
\begin{minipage}{0.53\textwidth}

\footnotesize

    \centering
    \resizebox{.99\textwidth}{!}{
    \setlength\tabcolsep{1.5pt}
    \begin{tabular}{lccccc}
    \toprule
        {\footnotesize Dataset} & \# {\footnotesize Images} & {\footnotesize Format} & {\footnotesize bboxes?} & \makecell[c]{\footnotesize free-\\viewing?} & \makecell[c]{\footnotesize human-\\centric?} \\
        \midrule
        VCR \cite{zellers2019recognition}            & 110K & QA     & $\checkmark$&& $\checkmark$\\
        VisualCOMET \cite{park2020visualcomet}       & 59K  & If/Then KB & $\checkmark$&& $\checkmark$\\
        Visual7W \cite{zhu2016visual7w}              & 47K  & QA     & $\checkmark$& partial &  \\
        Visual Madlibs \cite{yu2015visual}           & 11K  & FiTB   & $\checkmark$& partial  & $\checkmark$\\
        Abstract Scenes \cite{vedantam2015learning}  & 4.3K  & KB  & &&\\
        Why In Images \cite{pirsiavash2014inferring} & 792  & KB     &&& $\checkmark$ \\
        BD2BB \cite{pezzelle2020different}           & 3.2K & If/Then && $\checkmark$& $\checkmark$ \\
        FVQA \cite{wang2017fvqa}                     & 2.2K & QA+KB  &&&\\
        OK-VQA \cite{marino2019ok}                    & 14K & QA && $\checkmark$&\\
        KB-VQA \cite{wang2015explicit}               & 700 & QA & $\checkmark$&&\\
        \midrule
        \sherlock & 103K & clue/inference & $\checkmark$& $\checkmark$&\\
    \bottomrule
    \end{tabular}
    }
\end{minipage}
\hfill
\begin{minipage}{0.45\textwidth}
    
    \caption{Comparison between \sherlock and prior annotated corpora addressing visual abductive reasoning from static images. 
    \sherlock showcases a unique 
    data collection paradigm, leading to a rich variety of non-human centric 
    (i.e., not solely grounded in human references) visual abductive inferences. %
    }
    \label{tab:prior_dataset}
\end{minipage}
\end{table}

Using our corpus, we propose three complementary tasks that evaluate different aspects of machine capacity for visual abductive reasoning:
\begin{enumerate}[topsep=-3pt,itemsep=-1ex,partopsep=0pt,parsep=1ex]
\item \emph{\Retrievallongtitle:} given an image+region, the algorithm scores a large set of candidate \inferences and is rewarded for assigning a high score to the gold annotation.
\item \emph{\Localizationlongtitle:} the algorithm selects a bounding box within the image that provides the best evidence for a given \inference.
\item \emph{\Comparisonlongtitle:} the algorithm scores a small set of plausible inferences for a given image+region, and is rewarded for aligning its scores with human judgments over those sets.
\end{enumerate}

\noindent In our setup, a single model undertakes all of these tasks: we ask algorithms to score the plausibility of an \inference given an image and a bounding box contained within it.\footnote{We reserve generative evaluations (e.g., BLEU/CIDEr) for future work: shortcuts (e.g., outputting the technically correct ``this is a photo" for all inputs) make generation evaluation difficult in the abductive setting (see \S\ref{sec:sec_with_reference_to_generative_evaluations}). Nonetheless, generative \emph{models} can be evaluated in our setup; we experiment with one in \S \ref{sec:sec_with_results}.} We can directly compare models in their capacity to perform abductive reasoning, without relying on indirect generation evaluation metrics. %

Model predicted inferences are given in \fig~\ref{fig:first_examples}. The model is a fine-tuned CLIP \cite{radford2021learning} augmented to allow bounding boxes as input, enabling users to specify particular regions for the model to make abductive inferences about.
Our best model, a multitask version of CLIP \texttt{RN50x64}, outperforms strong baselines like UNITER \cite{chen2020uniter} and LXMERT \cite{tan2019lxmert} primarily because it 
pays specific attention to the correct input bounding box. We additionally show that 1) for all tasks, reasoning about the full context of the image (rather than just the region corresponding to the \clue) results in the best performance; 2) a text-only model cannot solve the \comparison task even when given oracle region descriptions; and 3) a multi-task model fit on both \clues/\inferences at training time performs best even when only \inferences are available at test time.

We foresee \sherlock as a difficult diagnostic benchmark for vision-and-language models. On our \comparison task, in terms of pairwise accuracy, our best model lags significantly below human agreement (headroom also exists for \retrieval and \localization). We release code, data, and models at \url{http://visualabduction.com/}.

\section{Related Work}

\label{sec:sec_with_motivation}

\noindent \textbf{Abductive reasoning.} 
Abduction, a form of everyday reasoning 
first framed by%
Peirce,
\cite{peirce1955philosophical,peirce1965pragmatism}; %
involves the creating of explanatory hypotheses%
based on limited evidence.
Humans use abduction to
reconcile seemingly disconnected observations to arrive at meaningful conclusions \cite{shank1998extraordinary} but readily retract in presence of new evidence \cite{aliseda2017logic}. In linguistics, %
abduction for communicated meaning (in an impoverished conversational context) is systematized through conversational maxims \cite{grice1975logic}. In images, \cite{berg2012understanding} show that different object types have different likelihoods of being mentioned in image captions (e.g., ``fireworks'' is always mentioned if depicted, but ``fabric'' is not), %
but that object type alone does not dictate salience for abductive inferences, e.g., %
a TV in a living room may not be as conceptually salient as a TV in a bar, which may
signal a particular type of bar.
Abductive reasoning has recently received attention in language processing tasks \cite{bhagavatula2019abductive,qin2020back,du2021learning,paul2021generating}, proof writing \cite{tafjord2020proofwriter}, and
discourse processing \cite{hobbs1993interpretation,ovchinnikova2011abductive}, etc.

\noindent \textbf{Beyond visual recognition.} %
Several tasks that go beyond image description/recognition have been proposed, including visual and analogical reasoning \cite{park2019robust,zhang2019raven,johnson2017clevr,antol2015VQA}, scene 
semantics \cite{johnson2015image}, commonsense interactions \cite{vedantam2015learning,pirsiavash2014inferring}, temporal/causal reasoning \cite{kim2022cosim,yi2019clevrer}, and perceived importance \cite{berg2012understanding}. Others have explored commonsense reasoning tasks posed over videos, which usually have more input available than a single frame \cite{tapaswi2016movieqa,jang2017tgif,lei2019tvqa+,zadeh2019social,garcia2020knowit,lei2020more,zhang2020learning,fang2020video2commonsense,liu2020violin,ignat2021whyact} (inter alia).

\noindent \textbf{Visual abductive reasoning.}
\sherlock builds upon prior grounded visual abductive reasoning efforts (Table~\ref{tab:prior_dataset}). Corpora like Visual Commonsense Reasoning (VCR) \cite{zellers2019recognition}, VisualCOMET \cite{park2020visualcomet}, and 
Visual7W \cite{zhu2016visual7w} are most similar to \sherlock in providing 
benchmarks for rationale-based inferences (i.e., the why and how).
But, \sherlock differs in format and content (\fig~\ref{tab:prior_work_comparison}). Instead of annotated QA pairs like in \cite{zhu2016visual7w,zellers2019recognition} where one option is definitively correct, free-text \clue/\inference pairs
allow for broader types of image descriptions, lending itself to 
softer and richer notions of reasoning (see \S \ref{sec:sec_with_task_setups})---\inferences are not definitively correct vs. incorrect, rather, they span a range of plausibility.
Deviating from the constrained, human-centric annotation of \cite{park2020visualcomet}, \sherlock \clue/\inference pairs support a broader range of topics
via our open-ended annotation paradigm (see \S\ref{sec:dataset-collection}). 
\sherlock's inferences can be grounded on any number of visual objects in an image, from figures central to the image (e.g., persons, animals, objects) to background cues %
(e.g., time, location, circumstances).

\section{\sherlock Corpus}
\label{sec:sherlock-corpus}

\noindent The \sherlock corpus contains a total of 363K abductive commonsense inferences grounded in 81K Visual Genome \cite{Krishna2016VisualGC} images (photographs from Flickr) and 22K Visual Commonsense Reasoning (VCR) \cite{zellers2019recognition} images (still-frames from movies). Images have an average of 3.5 \observations, each consisting of: 

\begin{itemize}[topsep=0pt,itemsep=-1ex,partopsep=0ex,parsep=1ex,leftmargin=*]
    \item \textbf{\clue}: an observable entity or object in the image, along with bounding box(es) specifying it (e.g., ``people wearing nametags''). 
    \item \textbf{\inference}: an abductive inference associated with the \clue; not immediately obvious from the image content (e.g., ``the people don't know each other''). 
\end{itemize}

\noindent Both \clues and \inferences are represented via free text in English; both have an average length of seven tokens; per \clue, there are a mean/median of 1.17/1.0 bounding boxes per \clue. We divide the 103K annotated images into a training/validation/test set of 90K/6.6K/6.6K. Further details are available in \S\ref{appendix:crowdsource}. %

\label{sec:dataset-collection}

\vspace{1mm} \noindent \textbf{Annotation process.} We crowdsource our dataset via Amazon Mechanical Turk (MTurk). 
For each data collection HIT, a manually qualified worker is given an image and prompted for 3 to 5 \observations. For each \observation, the worker is asked to write a \clue, highlight the regions in the image corresponding to the \clue, and 
write an \inference triggered by the \clue. To discourage
purely deductive reasoning, the workers are actively encouraged 
to think beyond the literally depicted scene, while %
working within 
real-world expectations.
Crowdworkers also self-report Likert ratings of confidence in the correctness of their abductive inferences along a scale of ``definitely" = 3/3, ``likely" = 2/3, and ``possibly" = 1/3.
The resulting inferences span this range
(31\%, 51\%, 18\%, respectively).
To validate corpus quality, we run a validation round 
for 17K \observations in which crowdworkers provide
ratings for \textit{acceptability} (is the annotation reasonable?),  \textit{bboxes} (are the boxes reasonably placed for the clue?), and \textit{interestingness} (how interesting is the annotation?). We find that 97.5\% of the \observations are acceptable with 98.3\% accurate box placement; and 71.9\% of inferences are found to be interesting. %

\subsection{Dataset Exploration} 
\noindent \sherlock's abductive inferences cover a wide variety of real world experiences from observations about unseen yet probable details of the image (e.g., ``smoke at an outdoor gathering'' $\rightarrow$ ``something is being grilled'') to elaborations on the expected social context (e.g., ``people wearing nametags'' $\rightarrow$ ``[they] don't know each other'').
Some \inferences are highly likely to be true 
(e.g., ``wet pavement'' $\rightarrow$ ``it has rained recently'');
others are less definitively verifiable, but nonetheless plausible (e.g., ``large trash containers'' $\rightarrow$ ``there is a business nearby''). Even the \inferences crowdworkers specify as 3/3 confident are almost always abductive, e.g., wet pavement strongly but not always indicate rain.
Through a rich array of natural observations,
\sherlock provides a tangible view into the abductive inferences people use on an everyday basis (more examples in \fig~\ref{fig:dataset_examples}).

\begin{figure}[t!]

\begin{minipage}{0.45\textwidth}
\includegraphics[width=\linewidth]{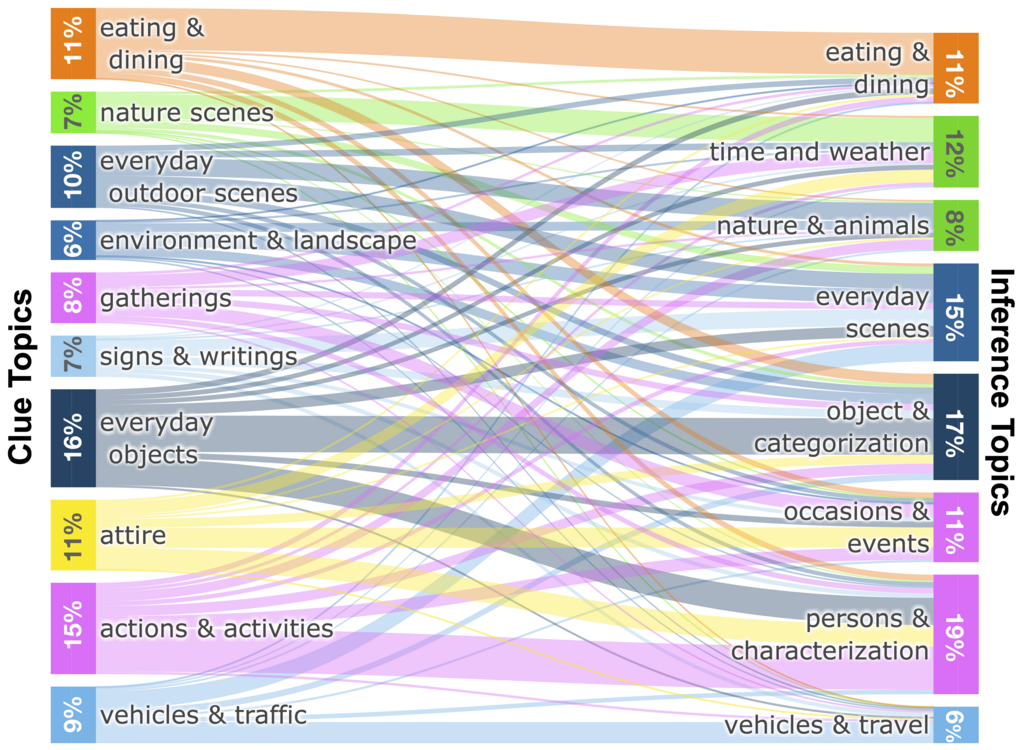}
\vspace{-1.5em}
\end{minipage}
\hfill
\begin{minipage}{0.5\textwidth}
\caption{
Overview of the topics represented in the \clues and \inferences in \sherlock. This analysis shows that $\sherlock$ covers a variety of 
topics
commonly accessible in the natural world. Color of the connections reflect the \clue topic.}

\label{fig:clues-to-inf-topics}
\end{minipage}
\end{figure}

\noindent \textbf{Assessing topic diversity.} To gauge the diversity of objects and situations represented in \sherlock, we run an LDA topic model \cite{blei2003latent} over the 
\observations.
The topics span a range of common everyday objects, entities, and situations (\fig~\ref{fig:clues-to-inf-topics}). \Inference topics associated with the \clues include within-category associations (e.g., ``baked potatoes on a ceramic plate'' $\rightarrow$ ``this
[is] a side dish'') and cross-category associations (e.g., ``a nametag'' (attire) $\rightarrow$ ``she works here'' (characterization)).
Many topics are not human centric;
compared to VCR/VisualCOMET in which 94\%/100\% of grounded references are to people. A manual analysis of 150 \clues reveals that only 36\% of \sherlock \observations are grounded on people.

\noindent \textbf{Intended use cases.} %
We manually examine of 250 randomly sampled \observations to better understand how annotators referenced protected characteristics (e.g., gender, color, nationality).
A majority of \inferences (243/250) are not directly about protected characteristics, though, a perceived gender is often made explicit via pronoun usage, e.g., ``she is running."
As an additional check, we pass 30K samples of our corpus through the Perspective API.\footnote{\url{https://www.perspectiveapi.com/}; November 2021 version. The API (which itself is imperfect and has biases \cite{hosseini2017deceiving,mitchell2019model,sap2019risk}) assigns toxicity value 0-1 for a given input text. Toxicity is defined as ``a rude, disrespectful, or unreasonable comment that is likely to make one leave a discussion.''} A manual examination of 150 cases marked as ``most toxic'' reveals mostly false positives (89\%), though 11\% of this sample do contain lewd content (mostly prompted by visual content in the R-rated VCR movies) or stigmas related to, e.g., gender and weight. See \S\ref{sec:sec_with_bias_investigation_appendix} for a more complete discussion.

\label{sec:sec_with_use_cases}
 While our analysis suggests that the relative magnitude of potentially offensive content is low in \sherlock, \emph{we still advocate against deployed use-cases that run the risk of perpetuating potential biases:} our aim is to \emph{study} abductive reasoning without endorsing the correctness or appropriateness of particular inferences. We foresee \sherlock as 1) a diagnostic corpus for measuring machine capacity for visual abductive reasoning; 2) a large-scale resource to study the types of inferences people may make about images; and 3) a potentially helpful resource for building tools that require understanding \emph{abductions} specifically, e.g., for detecting purposefully manipulative content posted online, it could be useful to specifically study what people \emph{might assume} about an image (rather than what is objectively correct; more details in Datasheet (\S\ref{sec:sec_with_datasheet}) \cite{gebru2018datasheets}).
\section{From Images to Abductive Inferences}

\label{sec:sec_with_retrieval}
\label{sec:sec_with_task_setups}
\begin{figure*}[t!]
\begin{center}
\begin{subfigure}[t]{0.32\textwidth}
\includegraphics[width=1\linewidth]{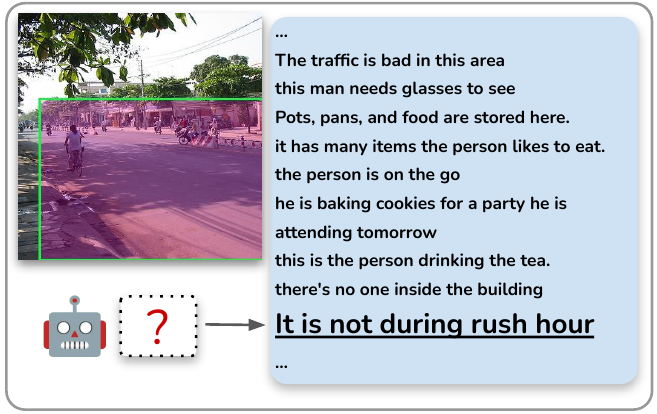}
\caption{\Retrievallongsent}
\label{fig:task-retrieval}
\end{subfigure}
\hfill
\begin{subfigure}[t]{0.32\textwidth}
\includegraphics[width=1\linewidth]{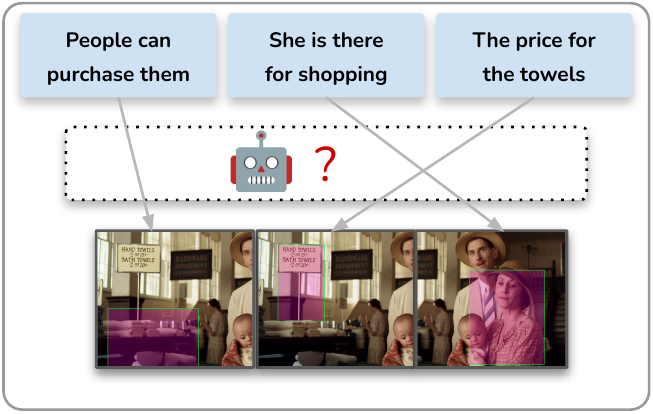}
\caption{\Localizationlongsent}
\label{fig:task-localization}
\end{subfigure}
\hfill
\begin{subfigure}[t]{0.32\textwidth}
\includegraphics[width=1\linewidth]{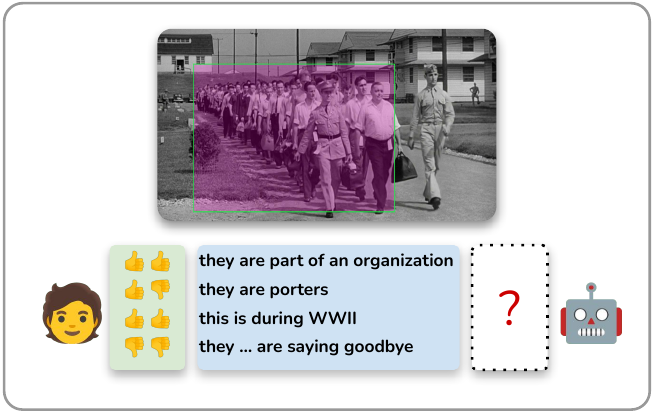}
\caption{\Comparisonlongsent}
\label{fig:task-comparison}
\end{subfigure}

\end{center}

\definecolor{likertcolor}{HTML}{6aa84f}
\definecolor{candidatecolor}{HTML}{3c78d8}
\caption{We pose three tasks over \sherlock: In \emph{\retrieval}, models are tasked with finding the ground-truth inference across a wide range of inferences, some much more plausible/relevant than others. In \emph{\localization}, models must align regions within the same image to several inferences written about that image. For \emph{\comparison}, we collect \textbf{\textcolor{likertcolor}{19K Likert ratings}} from human raters across \textbf{\textcolor{candidatecolor}{plausible candidates,}} and models are evaluated in their capacity to reconstruct human judgments across the candidates. Despite intrinsic subjectivity, headroom exists between human agreement and model performance, e.g., on the \emph{\comparison} task.}

\label{fig:likert_example}
\end{figure*}

\noindent We operationalize our corpus with three tasks, which we call \retrieval, \localization, and \comparison. Notationally, we say that an instance within the \sherlock corpus consists of an image $\boldsymbol{i}$, a region specified by $N$ bounding boxes $\boldsymbol{r} = \{\langle x_{1i}, x_{2i}, y_{1i}, y_{2i} \rangle\}_{i=1}^N$,\footnote{As discussed in \S\ref{sec:dataset-collection}, $N$ has a mean/median of 1.17/1.0 across the corpus.}  a \clue $\boldsymbol{c}$ corresponding to a literal description of $\boldsymbol{r}$'s contents, and an \inferencewithboldedf $\boldsymbol{f}$ that an annotator associated with $\boldsymbol{i}$, $\boldsymbol{r}$, and $\boldsymbol{c}$. We consider:

\begin{enumerate}[topsep=0pt,itemsep=-1ex,partopsep=1ex,parsep=1ex,leftmargin=*]
    \item \emph{\Retrievallongtitle:} For a given image/region pair $(\boldsymbol{i},\boldsymbol{r})$, how well can models select the ground-truth \inference $\boldsymbol{f}$ from a large set of candidates (\tildetext1K) covering a broad swath of the corpus?
    \item \emph{\Localizationlongtitle:} Given an image $\boldsymbol{i}$ and an \inference $\boldsymbol{f}$  written about an (unknown) region within the image, how well can models locate the proper region?
    \item \emph{\Comparisonlongtitle:} Given an image/region pair $(\boldsymbol{i},\boldsymbol{r})$ and a small set (\tildetext10) of relevant \inferences, can models predict how humans will rank their plausibility?
\end{enumerate}

\noindent Each task tests a complementary aspect of visual abductive reasoning (\fig~\ref{fig:likert_example}):
\retrieval tests across a broad range of \inferences, \localization tests within-images, and \comparison tests for correlation with human judgement.
Nonetheless, the same model can undertake all three tasks if it implements the following interface:

\label{sec:sec_with_interface}
\begin{mdframed}[style=TaskFrame]
\noindent \textbf{\sherlock Abductive Visual Reasoning Interface}
\begin{itemize}[leftmargin=*,topsep=0pt,itemsep=-1ex,partopsep=1ex,parsep=1ex]
\item \textbf{Input:} An image $\boldsymbol{i}$, a region $\boldsymbol{r}$ within $\boldsymbol{i}$, and a candidate \inference $\boldsymbol{f}$.
\item \textbf{Target:} A score $s$, where $s$ is proportional to the plausibility that $\boldsymbol{f}$ could be inferred from $\left( \boldsymbol{i}, \boldsymbol{r} \right)$.
\end{itemize}
\end{mdframed}

\noindent That is, we assume a model $m: \left( \boldsymbol{i}, \boldsymbol{r}, \boldsymbol{f} \right) \rightarrow \mathbb{R}$ that scores \inference $\boldsymbol{f}$'s plausibility for $\left( \boldsymbol{i}, \boldsymbol{r} \right)$. Notably, the interface takes as input \inferences, but not \clues: our intent is to focus evaluation on abductive reasoning, rather than the distinct setting of literal referring expressions.\footnote{In \S\ref{sec:sec_with_clue_results}, for completeness, we give results on the \retrieval and \localization setups, but testing on \clues instead.} \Clues can be used for training $m$; as we will see in \S\ref{sec:sec_with_models} our best performing model, in fact, does use \clues at training time.

\subsection{\Retrievallongtitle}
\label{sec:model_retrieval}

\noindent For \retrieval evaluation, at test time, we are given an $\left(\boldsymbol{i}, \boldsymbol{r} \right)$ pair, and a large (\tildetext1K)\footnote{Our validation/test sets contain about 23K \inferences. For efficiency we randomly split into 23 equal sized chunks of about 1K \inferences, and report \retrieval averaged over the resulting splits.} set of candidate \inferences $\boldsymbol{f} \in \boldsymbol{F}$, only one of which was written by an annotator for $\left( \boldsymbol{i}, \boldsymbol{r} \right)$; the others are randomly sampled from the corpus. %
In the $im \rightarrow txt$ direction, we compute the mean rank of the true item (lower=better) and $P@1$ (higher=better); in the $txt \rightarrow im$ direction, we report mean rank (lower=better). %

\subsection{\Localizationlongtitle}
\label{sec:model_localization}

\noindent \Localization assesses a model's capacity select a regions within an image that most directly supports a given \inference. %
Following prior work on literal referring expression localization \cite{krahmer2012computational,kazemzadeh2014referitgame,yu2016modeling} (inter alia), we experiment in two settings: 1) we are given all the ground-truth bounding boxes for an image, and 2) we are given only automatic bounding box proposals from an object detection model.

\noindent \textbf{GT bounding boxes.} We assume an image $\boldsymbol{i}$, the set of 3+ \inferences $\boldsymbol{F}$ written for that image, and the (unaligned) set of regions $\boldsymbol{R}$ corresponding to $\boldsymbol{F}$. The model must produce a one-to-one assignment of $\boldsymbol{F}$ to $\boldsymbol{R}$ in the context of $\boldsymbol{i}$. In practice, we score all possible $\boldsymbol{F} \times \boldsymbol{R}$ pairs via the abductive visual reasoning interface, and then compute the maximum linear assignment \cite{kuhn1955hungarian} using \href{https://github.com/src-d/lapjv}{lapjv}'s implementation of \cite{jonker1987shortest}. The evaluation metric is the accuracy of this assignment, averaged over all images. To quantify an upper bound, a human rater performed the assignment for 101 images, achieving an average accuracy of 92.3\%.

\noindent \textbf{Auto bounding boxes.} We compute 100 bounding box proposals per image by applying Faster-RCNN \cite{ren2015faster} with a ResNeXt101 \cite{xie2017aggregated} backbone trained on Visual Genome to all the images in our corpus. Given an image $\boldsymbol{i}$ and an \inference $\boldsymbol{f}$ that was written about the image, we score all 100 bounding box proposals independently and take the highest scoring one as the prediction. We count a prediction as correct if it has IoU $>0.5$ with a true bounding box that corresponds to that \inference,\footnote{Since the annotators were able to specify multiple bounding boxes per \observation, we count a match to any of the labeled bounding boxes.} and incorrect otherwise.\footnote{A small number of images do not have a ResNeXt bounding box with IoU$>0.5$ with any ground truth bounding box: in \S \ref{sec:sec_with_results}, we show that most instances (96.2\%) are solvable with this setup.}

\newcommand{\nodata}{-}
\begin{table*}[t]
\centering
\resizebox{.99\textwidth}{!}{
\begin{tabular}{lccccc}
  & \multicolumn{3}{c}{\emph{\Retrieval}} & \emph{\Localization} & %
  \emph{\Comparison}
  \\
  \cmidrule(lr){2-4}\cmidrule(lr){5-5}\cmidrule(lr){6-6}
  & im $\rightarrow$ txt ($\downarrow$) & txt $\rightarrow$ im ($\downarrow$) & $P@1_{im \rightarrow txt}$ ($\uparrow$) & GT-Box/Auto-Box ($\uparrow$) & Val/Test Human Acc ($\uparrow$) \\
  \midrule
  Random & 495.4 & 495.4 & 0.1 & 30.0/7.9 & 1.1/-0.6\\
  Bbox Position/Size & 257.5 & 262.7 & 1.3 & 57.3/18.8 & 5.5/1.4 \\
  \midrule
  LXMERT  & 51.1 & 48.8 & 14.9 & 69.5/30.3 & 18.6/21.1 \\
  UNITER Base & 40.4 & 40.0 & 19.8 &  73.0/33.3 & 20.0/22.9\\
  CLIP \texttt{ViT-B/16} &  19.9 & 21.6 & 30.6 &  85.3/38.6 & 20.1/21.3\\
  CLIP \texttt{RN50x16} & 19.3 & 20.8 & 31.0 &  85.7/38.7 & 21.6/23.7\\
  CLIP \texttt{RN50x64} & 19.3 & 19.7 & 31.8 &  86.6/39.5 & 25.1/26.0 \\
  \rotatebox[origin=c]{180}{$\Lsh$} + multitask \clue learning & \textbf{16.4} & \textbf{17.7} & \textbf{33.4} &  \textbf{87.2}/\textbf{40.6} & \textbf{26.6}/\textbf{27.1}\\
  \midrule
  \underline{Human} + (Upper Bound) & \nodata& \nodata & \nodata & \underline{92.3}/(96.2) &
   \underline{42.3/42.3} \\
   \bottomrule
\end{tabular}

}

\caption{Test results for all models across all three tasks. CLIP \texttt{RN50x64} outperforms all models in all setups, but significant headroom exists, e.g., on \Comparison between the model and human agreement.
}
\label{tab:retrieval_results}
\end{table*}

\subsection{\Comparisonlongtitle}

\label{sec:model_reasoning}

\noindent We assess model capacity to make fine-grained assessments given a set of plausible inferences. For example, in \fig~\ref{fig:task-comparison} (depicting a group of men marching and carrying bags), human raters are \textit{likely} to 
say that they are military men and that the photo was taken during WWII, and \textit{unlikely} to see them as porters despite them carrying bags.
Our evaluation assumes that a performant model's predictions should correlate with the (average) relative judgments made by humans, and we seek to construct a corpus that supports evaluation of such reasoning.

\noindent \textbf{Constructing sets of plausible \inferences.} %
We use a performant model checkpoint fine-tuned for the Sherlock tasks\footnote{Specifically, a \texttt{CLIP RN50x16} checkpoint that achieves strong validation \retrieval performance (comparable to the checkpoint of the reported test results
in \S \ref{sec:sec_with_results}); model details in \S \ref{sec:sec_with_models}.} %
to compute the similarity score between all $(\boldsymbol{i}, \boldsymbol{r}, \boldsymbol{f})$ triples in the validation/test sets. 
Next, we perform several filtering steps: 1) we only consider pairs where the negative \inference received a higher score than the ground-truth according to the model; 2) we perform soft text deduplication to downsample \inferences that are semantically similar; and 3) we perform hard text deduplication, only allowing \inferences to appear verbatim 3x times. Then, through an iterative process, we uniquely sample a diverse set of 10 \inferences per $\left( \boldsymbol{i}, \boldsymbol{r} \right)$ that meet these filtering criteria. 
This results in a set of 10 plausible \inference candidates for each of 485/472 validation/test images. More details are in \S\ref{sec:sec_w_human_evaluation_details}. In a retrieval sense, these plausible \inferences can be viewed as ``hard negatives:" i.e., none are the gold annotated \inference, but a strong model nonetheless rates them as plausible.

\noindent \textbf{Human rating of plausible \inferences.} Using MTurk%
, we collect two annotations of each candidate \inference on a three-point Likert scale ranging from 1 (bad: ``irrelevant"/``verifiably incorrect") to 3 (good: ``statement is probably true; the highlighted region supports it.").  We collect 19K annotations in total (see \S\ref{sec:sec_w_human_evaluation_details} for full details).
Because abductive reasoning involves subjectivity and uncertainty, we expect some amount of intrinsic disagreement between raters.\footnote{In \S \ref{sec:sec_with_results}, we show that models achieve significantly less correlation compared to human agreement.}
We measure model correlation with human judgments on this set via pairwise accuracy. For each image, for all pairs of candidates that are rated differently on the Likert scale, the model gets an accuracy point if it orders them consistently with the human rater's ordering. Ties are broken randomly but consistently across all models. For readability, we subtract the accuracy of a random model (50\%) and multiply by two to form the final accuracy metric.
\section{Methods and Experiments}

\label{sec:sec_with_models}

\noindent \textbf{Training objective.} To support the interface described in \S \ref{sec:sec_with_interface},
we train models $m: \left( \boldsymbol{i}, \boldsymbol{r}, \boldsymbol{f} \right) \rightarrow \mathbb{R}$ that score \inference $\boldsymbol{f}$'s plausibility for $\left( \boldsymbol{i}, \boldsymbol{r} \right)$. We experiment with several different V+L backbones as detailed below; for each, we train by optimizing model parameters to score truly corresponding $\left( \boldsymbol{i}, \boldsymbol{r}, \boldsymbol{f} \right)$ triples more highly than negatively sampled $\left( \boldsymbol{i}, \boldsymbol{r}, \boldsymbol{f}_{fake} \right)$
triples.

\noindent \textbf{LXMERT} \cite{tan2019lxmert} is a vision+language transformer \cite{vaswani2017attention} model pre-trained on Visual Genome \cite{Krishna2016VisualGC} and MSCOCO \cite{Lin2014MicrosoftCC}. The model is composed of three transformer encoders \cite{vaswani2017attention}: an object-relationship encoder (which takes in ROI features+locations with a max of 36, following \cite{Anderson2017BottomUpAT}), %
a language encoder that processes word tokens,
and a cross modality encoder. To provide region information $r$, we calculate the ROI feature of $r$ and always place it in the first object token to the visual encoder (this is a common practice for, e.g., the VCR dataset \cite{zellers2019recognition}). We follow \cite{chen2020uniter} to train the model in ``image-text retrieval" mode by maximizing the margin $m=.2$ between the cosine similarity scores of positive triple $\left( \boldsymbol{i}, \boldsymbol{r}, \boldsymbol{f} \right)$ and two negative triples $\left( \boldsymbol{i}, \boldsymbol{r}, \boldsymbol{f}_{fake} \right)$ and $\left( \boldsymbol{i}_{fake}, \boldsymbol{r}_{fake}, \boldsymbol{f} \right)$ through triplet loss.

\noindent \textbf{UNITER} \cite{chen2020uniter} %
consists of a single, unified transformer that takes in image and text embeddings. We experiment with the Base version pre-trained on MSCOCO \cite{Lin2014MicrosoftCC}, Visual Genome \cite{Krishna2016VisualGC}, Conceptual Captions \cite{Sharma2018ConceptualCA}, and SBU Captions \cite{Ordonez2011Im2TextDI}. 
We apply the same strategy of region-of-reference-first passing %
and train with the same triplet loss following \cite{chen2020uniter}.

\noindent \textbf{CLIP.} 
We finetune
the \texttt{ViT-B/16}, \texttt{RN50x16}, and \texttt{RN50x50} versions of CLIP \cite{radford2021learning}.
Text is represented via a 12-layer text transformer. For \texttt{ViT-B/16}, images are represented by a 12-layer vision transformer \cite{dosovitskiy2020image}, whereas for \texttt{RN50x16}/\texttt{RN50x64}, images are represented by EfficientNet-scaled ResNet50 \cite{he2016deep,tan2019efficientnet}.

\begin{figure*}[t]
    \centering
    \begin{subfigure}[b]{0.5\textwidth}
     \resizebox{\linewidth}{!}{
      
\begin{tabular}[b]{clcc}
\toprule
  & & $P@1$ ($\uparrow$) & Val/Test Human ($\uparrow$) \\
  \midrule
  & CLIP \texttt{ViT-B/16} & 30.5 & 20.1/21.2 \\
  \midrule
  \parbox[t]{1.5mm}{\multirow{5}{*}{\rotatebox[origin=c]{90}{input}}}
  
  &\rotatebox[origin=c]{180}{$\Lsh$} Position only & 1.3 & 5.5/1.4 \\
  &\rotatebox[origin=c]{180}{$\Lsh$} No Region & 18.1 & 16.8/19.0 \\
  &\rotatebox[origin=c]{180}{$\Lsh$} No Context & 24.8 & 18.1/17.8 \\ 
  &\rotatebox[origin=c]{180}{$\Lsh$} Only context & 18.9 & 17.4/16.3 \\
  &\rotatebox[origin=c]{180}{$\Lsh$} Trained w/ only Clues &  23.0 & 16.2/19.7 \\
  \midrule
  \parbox[t]{1.5mm}{\multirow{3}{*}{\rotatebox[origin=c]{90}{model}}}

  &\rotatebox[origin=c]{180}{$\Lsh$} Crop no Widescreen & 27.8 & 23.1/21.8 \\
  &\rotatebox[origin=c]{180}{$\Lsh$} Resize no Widescreen & 27.7 & 19.4/20.6 \\
  &\rotatebox[origin=c]{180}{$\Lsh$} Zero shot w/ prompt & 12.0 & 10.0/9.5 \\
\bottomrule
\end{tabular}

     }
     \caption{}
     \label{fig:ablations_main_results}
    \end{subfigure}
    \hfill
    \begin{subfigure}[b]{0.49\textwidth}
     \centering
     \includegraphics[width=.85\linewidth]{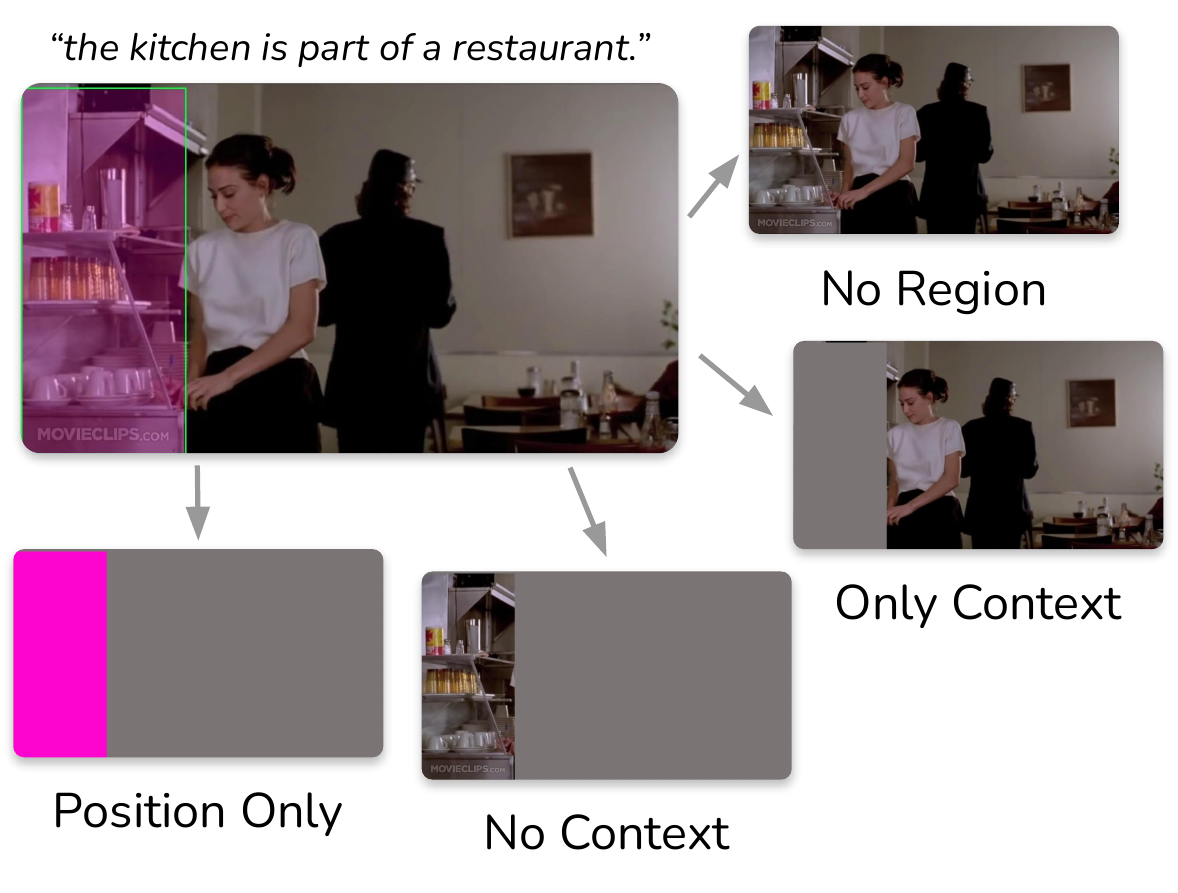}
     \caption{}
     \label{fig:ablations_main_ex}
    \end{subfigure}
  \caption{We perform ablations by varying the input data, top (a), and the modeling components, bottom (a). Figure (b) depicts our image input ablations, which are conducted by drawing in pixel-space directly, following \cite{zellers2021merlot}. Having no context may make it difficult to situate the scene more broadly; here: neatly stacked cups could be in a bar, a hotel, a store, etc. Access only the context of the dining room is also insufficient. For modeling, bottom (a), cropping/resizing decreases performance on \retrieval ($P@1$), but not \comparison (Val/Test Human).}
  \label{fig:ablations_main}
\end{figure*}

We modify CLIP to incorporate the bounding box as input. Inspired by a similar process from \cite{zellers2021merlot,yao2021cpt}, to pass a region to CLIP, we simply draw a bounding box on an image in pixel space---we use a green-bordered / opaque purple box as depicted in \fig~\ref{fig:ablations_main_ex} (early experiments proved this more effective than modifying CLIP's architecture).
To enable CLIP to process the widescreen images of
VCR, we apply it twice to the input using overlapping square regions, i.e., graphically, like this: $[_1 [_2 ]_1 ]_2$, and average the resulting embeddings. We fine-tune using InfoNCE \cite{sohn2016improved,oord2018representation}. We sample a batch of truly corresponding $\left( \boldsymbol{i}, \boldsymbol{r}, \boldsymbol{f} \right)$ triples, render the regions $r$ in their corresponding images, and then construct all possible negative $\left( \boldsymbol{i}, \boldsymbol{r}, \boldsymbol{f}_{fake} \right)$ triples in the batch by aligning each \inference to each $\left( \boldsymbol{i}, \boldsymbol{r} \right)$. %
We use the biggest minibatch size possible using 8 GPUs with 48GB of memory each: 64, 200, and 512 for \texttt{RN50x64}, \texttt{RN50x16}, and \texttt{ViT-B/16}, respectively.

\noindent \textbf{Multitask learning.} All models thus far only utilize \inferences at training time. We experiment with a multitask learning setup using CLIP that additionally trains with \clues. In addition to training using our abductive reasoning objective, i.e., InfoNCE on \inferences, we mix in an additional referring expression objective, i.e., InfoNCE on \clues. Evaluation remains the same: at test time, we do not assume access to \clues. At training time, for each observation, half the time we sample an \inference (to form $\left( \boldsymbol{i}, \boldsymbol{r}, \boldsymbol{f} \right)$, and half the time we sample a \clue (to form $\left( \boldsymbol{i}, \boldsymbol{r}, \boldsymbol{c} \right)$). The \clue/\inference mixed batch of examples is then handed to CLIP, and a gradient update is made with InfoNCE as usual. To enable to model to differentiate between \clues/\inferences, we prefix the texts with \texttt{clue:}/\texttt{inference:}, respectively.

\noindent \textbf{Baselines.} In addition to a random baseline, we consider a content-free version of our CLIP \texttt{ViT-B/16} model that is given only the position/size of each bounding box. In place of the image, we pass a mean pixel value across the entire image and draw the bounding box on the image using an opaque pink box (see \S \ref{ssec:section_with_ablations}). 

\subsection{Results}

\label{sec:sec_with_results}

\noindent Table~\ref{tab:retrieval_results} contains results for all the tasks: In all cases, our CLIP-based models perform best, with \texttt{RN50x64} outperforming its smaller counterparts. Incorporating the multitask objective pushes performance further. While CLIP performs the best, UNITER is more competitive on \comparison and less competitive on \retrieval and \localization. We speculate this has to do with the nature of each task: 
\retrieval requires models to reason about many incorrect examples, whereas, the \inferences in the \comparison task are usually relevant to the objects in the scene.
In \S\ref{sec:sec_with_batch_size_ablation}, we provide ablations that demonstrate CLIP models outperform UNITER even when trained with a smaller batch size. Compared to human agreement on \comparison, our best model only gets 65\% of the way there (27\% vs. 42 \%).

\subsection{Ablations}
\label{ssec:section_with_ablations}

\noindent We perform data and model ablations on CLIP \texttt{ViT-B/16}. Results are in \fig~\ref{fig:ablations_main}.

\noindent \textbf{Input ablations.} Each part of our visual input is important. Aside from the position only model, the biggest drop-off in performance results from not passing the region as input to CLIP, e.g., $P@1$ for $im \rightarrow txt$ retrieval nearly halves, dropping from 31 to 18, suggesting that
CLIP
relies on the local region information to reason about the image.
Removing the region's content (``Only Context") unsurprisingly hurts performance, but so does removing the surrounding context (``No Context"). That is, the model performs the best when it can reason about the \clue and its full visual context jointly. On the text side, we trained a model with only \clues; \retrieval and \comparison performance both drop, which suggests that \clues and \inferences carry different information (additional results in \S\ref{sec:sec_with_clue_results}).

\begin{figure*}[t!]

 \begin{minipage}{.36\linewidth}
    \centering
    \includegraphics[width=.8\linewidth]{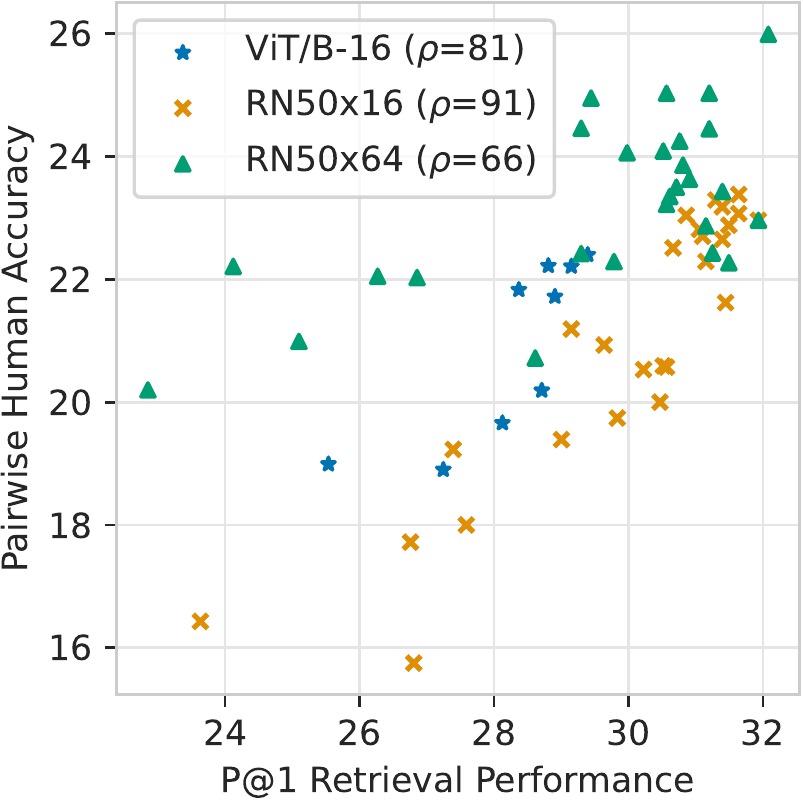}
    \captionof{figure}{Validation \retrieval perf. ($P@1$) vs. \comparison acc. for CLIP checkpoints. 
    }
    \label{fig:retrieval_vs_human}
 \end{minipage}
 \begin{minipage}{.57\linewidth}
    \centering
    \includegraphics[width=\linewidth]{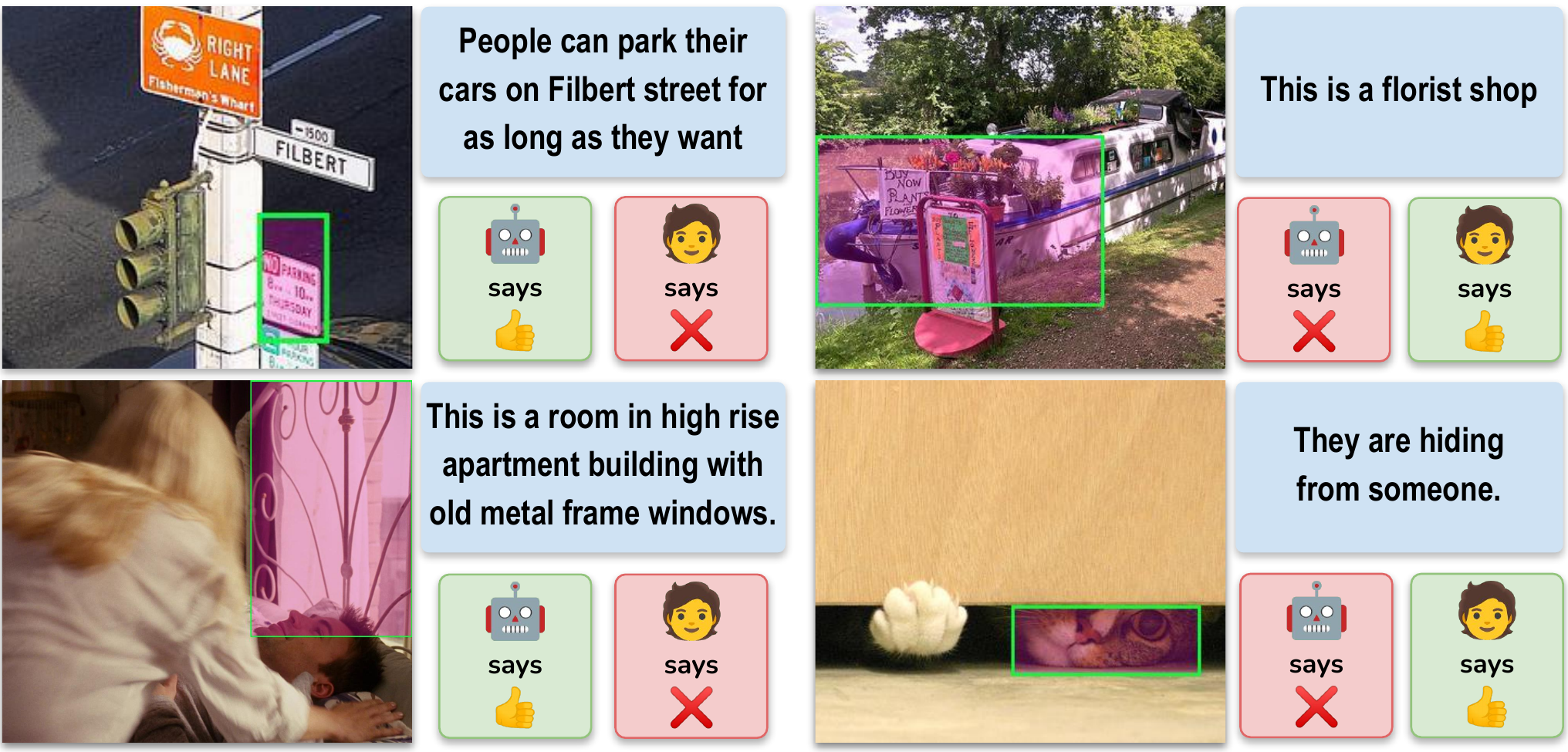}
    \captionof{figure}{Error analysis: examples of false positives and false negatives predicted by our model on the \comparison task's validation set.}
    \label{fig:error_examples}
  \end{minipage}

\end{figure*}

\noindent \textbf{Model ablations.} We considered two alternate image processing configurations.
Instead of doing two CLIP passes per image to facilitate widescreen processing (%
\S\ref{sec:sec_with_models}), we consider (i) center cropping and (ii) pad-and-resizing. Both take less computation, but provide less information to the model. Cropping removes the sides of images, whereas pad-and-resize lowers the resolution significantly. The bottom half of the table in \fig~\ref{fig:ablations_main_results} reports the results: both configurations lower performance on \retrieval tasks, but there's less impact for \comparison.

\noindent \textbf{Better \retrieval $\rightarrow$ better \comparison.} In \fig~\ref{fig:retrieval_vs_human}, we observe a high correlation between the \retrieval performance of our (single-task) CLIP model checkpoints ($P@1$) and the \comparison human accuracy for the \comparison task. %
For the smaller
\texttt{RN50x16} and \texttt{ViT-B/16} models, this effect cannot simply be explained by training time; for \texttt{RN50x16}, pearson corr. between training steps and \comparison performance is 81, whereas, the correlation between $P@1$ and \comparison performance is 91.  Overall, it's plausible that a model with higher precision at retrieval could help further bridge the gap on the \comparison task.

\noindent \textbf{Oracle text-only models are insufficient.}
One potential concern with our setup is that \clues may map one-to-one onto \inferences, e.g., if all soccer balls in our corpus were mapped onto ``the owner plays soccer" (and vice versa). We compare to an oracle baseline that makes this pessimistic assumption (complementing
our ``No Context" ablation, which provides a comparable context-free \emph{visual} reference to the \clue). 
We give the model oracle access to the ground-truth clues. Following \cite{bhagavatula2019abductive}, we use T5-Large v1.1 \cite{raffel2019exploring} to map \clues to \inferences with no access to the image by fitting $P(\text{\inference} | \text{clue})$ in a sequence-to-sequence fashion; training details are in \S\ref{sec:sec_with_modeling_details}.
The resulting text-only \clue $\rightarrow$ \inference model, when given the \clue \emph{``chipped paint and rusted umbrella poles"}, estimates likely \inferences, for example:
\emph{``the area is in a disrepair"},
\emph{``the city does not care about its infrastructure."}, etc. 
The text-only oracle under-performs vs. CLIP \emph{despite the fact that, unlike CLIP, it's given the ground-truth \clue}: on \comparison, it achieves 22.8/19.3 val/test accuracy; significantly lower than 26.6/27.1 that our best vision+language model achieves. This is probably because
global scene context cannot be fully summarized via a local referring expression. In the prior \emph{``chipped paint and rusted umbrella poles"} example, the true inference, \emph{``this beach furniture does not get put inside at night"}, requires additional visual context beyond the \clue---chipped paint and a rusty umbrella alone may not provide enough context to infer that this furniture is \emph{beach} furniture.

\subsection{Error Analysis}

We conduct a quantitative error analysis of multitask CLIP \texttt{RN50x64} for the \comparison task. We select  
340 validation images with highest human agreement, and split images into two groups: one where the model performed above average, and one where the model performed below average. We attempt to predict into which group an image will fall using logistic regression in 5-fold cross-validation. Overall, errors are difficult to predict. Surface level image/text features of the images/\inferences are not very predictive of errors: relative to a 50\% ROC AUC baseline, CLIP \texttt{ViT-B/16} image features achieve 55\%, whereas the mean SentenceBERT \cite{reimers-2019-sentence-bert} embedding of the \inference achieves 54\%. While not available \emph{a priori}, more predictive than content features of model errors are human Likert ratings: a single-feature mean human agreement model achieves 57\% AUC, (more human agreement = better model performance). %

\fig~\ref{fig:error_examples} gives qualitative examples of false positives/negatives. The types of abductive reasoning the model falls short on are diverse. In the boat example, the model fails to notice that a florist has set up shop on a ship deck; in the window example, the model misinterprets the bars over the windows as being \emph{outside} the building versus inside and attached to a bed-frame. The model is capable of reading some simple signs, but, as highlighted by \cite{mishra2019ocr}, reasoning about the semantics of written text placed in images remains a challenge, e.g., a ``no parking" sign is misidentified as an ``okay to park" sign. Overall: the difficult-to-categorize nature of these examples suggests that the \sherlock corpus makes for difficult benchmark for visual abductive reasoning.

\section{Conclusion}
\label{sec:sec_with_reference_to_generative_evaluations}
We introduce \sherlock, a corpus of visual abductive reasoning containing 363K \clue/\inference \observations across 103K images. Our work complements existing abductive reasoning corpora, both in format (free-viewing, free-text) and in diversity (not human-centric). Our work not only provides a challenging vision+language benchmark, but also, we hope it can serve as a resource for studying visual abductive reasoning more broadly. Future work includes:
\begin{enumerate}[leftmargin=*,topsep=0pt,itemsep=-1ex,partopsep=1ex,parsep=1ex]
\item Salience: in \sherlock, annotators specify salient \clues; how/why does salience differ from other free-viewing setups, like image captioning?
\item Ambiguity: 
when/why do people (justifiably) come to different conclusions?
\item Generative evaluation metrics: generation evaluation in abductive setting, i.e., without definitive notions of correctness, remains a challenge.
\end{enumerate}

\paragraph{Acknowledgments.} This work was funded by DARPA MCS program through NIWC Pacific (N66001-19-2-4031), the DARPA SemaFor program, and the Allen Institute for AI. AR was additionally in part supported by the DARPA PTG program, as well as BAIR's industrial alliance program. We additionally thank the UC Berkeley Semafor group for the helpful discussions and feedback.

\clearpage

\bibliographystyle{splncs04}
\bibliography{egbib}

\appendix
\clearpage
\section*{Supplementary Material}
\section{\sherlock Data Collection and Evaluation}
\label{appendix:crowdsource}

\begin{figure*}

\begin{center}
\includegraphics[width=0.9\linewidth]{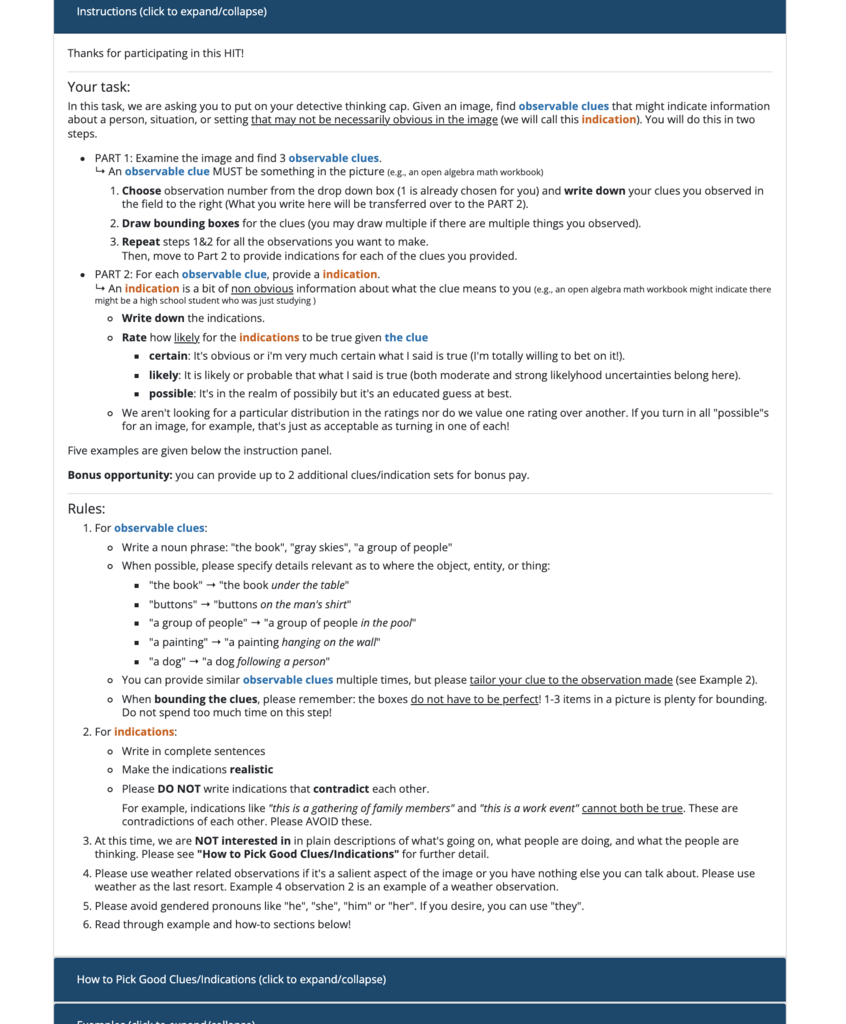}
\end{center}

\caption{
Instructions for \sherlock data collection HIT.
}

\label{fig:sample_hit_instr}
\end{figure*}

The dataset was collected during the month of February of 2021. The data collected is in English and HITs were open to workers originating from US, Canada, Great Britain and Australia. We target for a worker payment rate of $\$15$/hour for all our HITs. For data collection and qualifications, average pay for the workers came to $\$16$-$\$20$ with median workers being compensated $\$12$/hour. We hash Worker IDs to preserve anonymity. A sample of data collection HIT is shown in \fig~\ref{fig:sample_hit} (with instructions shown in \fig~\ref{fig:sample_hit_instr}).

\subsection{Qualification of Workers}

As a means for ensuring high quality annotations, 266 workers were manually selected through a qualification and training rounds. The workers were presented with three images and asked to provide three \observations per image. Each of the worker responses were manually evaluated. A total 297 workers submitting 8 reasonable \observations out of of 9 were qualified for training. 

\begin{figure*}

\begin{center}
\includegraphics[width=0.9\linewidth]{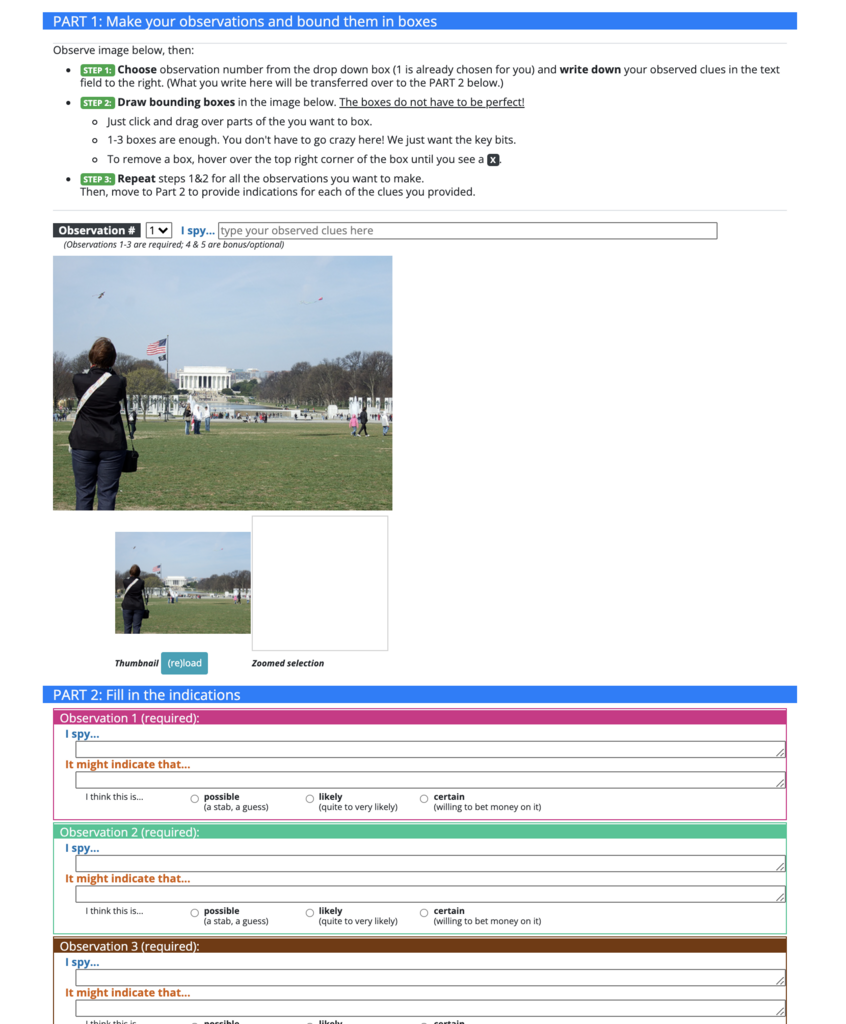}
\end{center}

\caption{
Template setup for \sherlock data collection HIT. Instructions are shown in Figure \ref{fig:sample_hit_instr}
}

\label{fig:sample_hit}
\end{figure*}

The process of creating bounding boxes and linking these boxes to the \observations was complex enough to necessitate a training stage. For the training round, qualified workers were given a standard data collection hit (\fig~\ref{fig:sample_hit}) at a higher pay to account for the time expected for them to learn the process. An additional training round was encouraged for a small pool of workers to ensure all workers were on the page with regards to the instructions and the mechanism of the hit. 266 workers worked on and completed the training (remaining 31 did not return for the training round). In this paper, we use the term \textit{qualified workers} to refer to the workers who have completed both the qualification and training round.

\subsection{Data Collection}
\label{appendix:data-collection}

\begin{figure*}[t!]

\begin{center}
\includegraphics[width=0.93\linewidth]{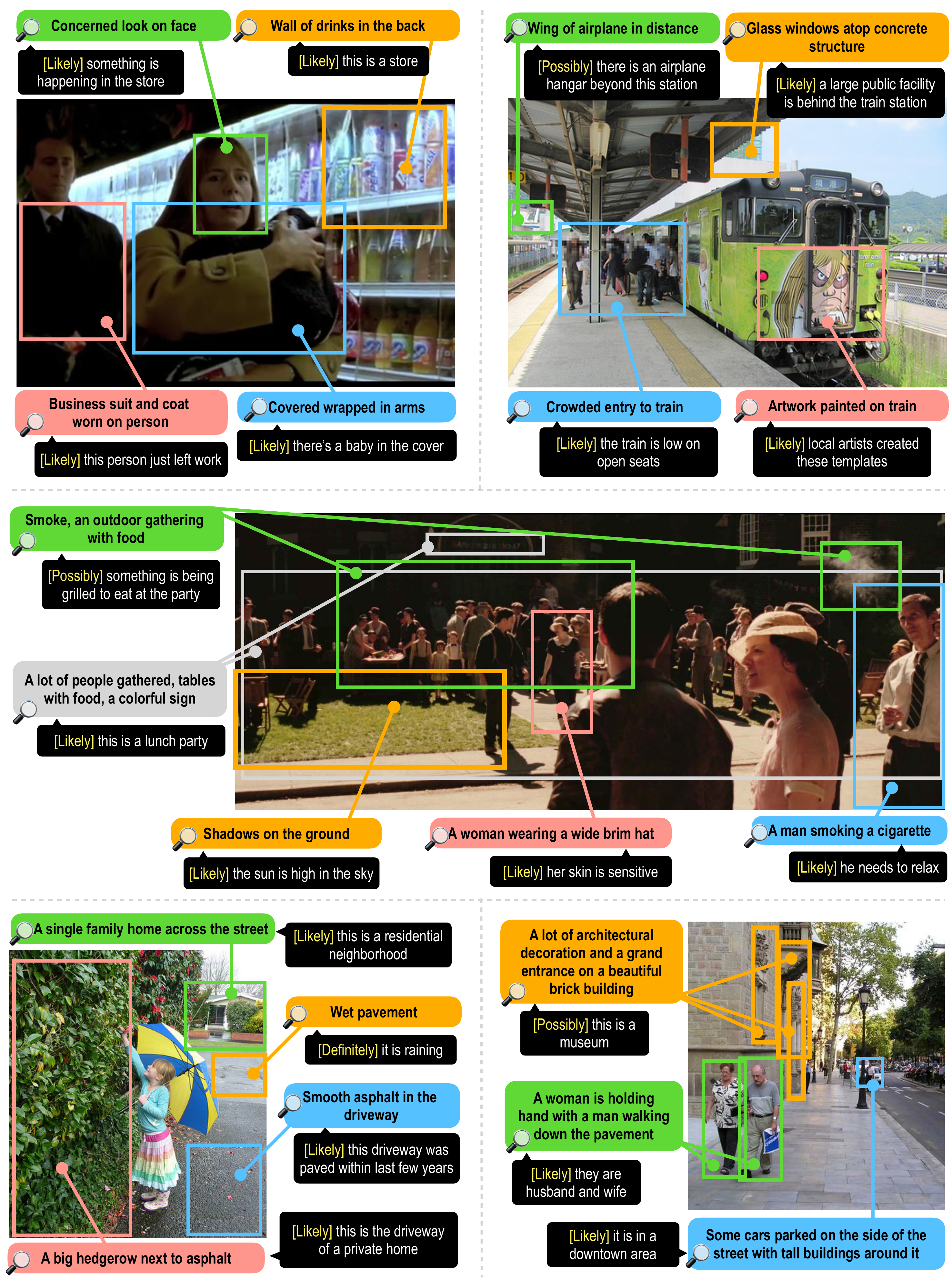}
\end{center}
\caption{
Examples of clues and inference pair annotations in \sherlock{} over images from Visual Genome and VCR. For each \observation, an \inference (speech bubble) is grounded in a concrete \clue (color bubble) present in an image. \confidence (in the order of decreasing confidence: ``Definitely'' $>$ ``Likely'' $>$ ``Possibly'') for each \inference is shown in yellow.}
\label{fig:dataset_examples}
\end{figure*}

As described in \S\ref{sec:sherlock-corpus}, we collected a total of 363K \observations which consist of a \clue and \inference. Further examples of annotations are shown in \fig~\ref{fig:dataset_examples}. 

\noindent \textbf{Image sourcing.} %
For VCR images, we use the subset also annotated by VisualCOMET \cite{park2020visualcomet}; we limit our selection to images that contain at least 3 unique entities (persons or objects).
For Visual Genome, during early annotation rounds, crowdworkers shared that particular classes of images were common and less interesting (e.g., grazing zebras, sheep in pastures). In response, 
we performed a semantic de-duplication step by hierarchical clustering into 80K clusters of extracted CLIP \texttt{ViT-B/32} features~\cite{radford2021learning} and sample a single image from each resulting cluster. 
We annotate 103K images in total, and divide them into a training/validation/test set of 90K/6.6K/6.6K, aligned with the community standard splits for these corpora.

\paragraph{Bounding boxes.} For each \clue in an \observation, the workers were asked to draw one or more bounding boxes around image regions relevant to the \clue. For example, for the \clue ``a lot of architectural decorations'' given for the lower right image in \fig~\ref{fig:dataset_examples}, the worker chose box each of the architectural features separately in their own bounding box. While it was not strictly enforced, we encouraged the workers to keep to a maximum of $3$ bounding boxes per \clue, with allowance for more if necessitated by the image and the \observation, based on worker's individual discretion.

\subsection{Corpus Validation}
\label{appendix:corpus_validation}

\begin{figure*}

\begin{center}
\includegraphics[width=0.9\linewidth]{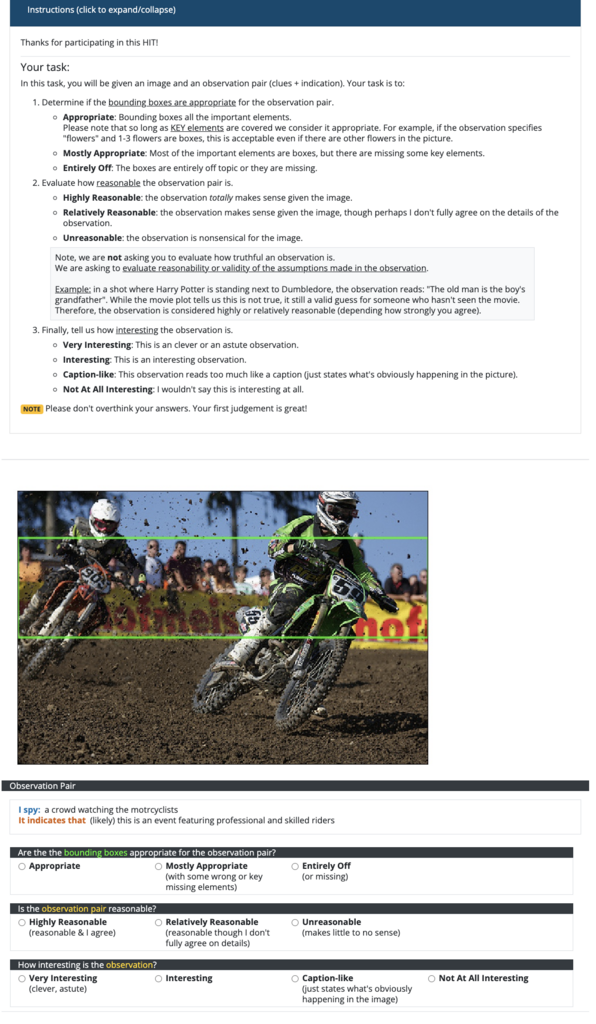}
\end{center}

\caption{
Instructions and template setup for \sherlock data validation HIT. 
}

\label{fig:sample_val_hit}
\end{figure*}

To verify the quality of annotation, we run a validation over 17K \observations.  For each \observation, we present three independent crowdworkers with its associated image and its annotation: the \clue with its corresponding region bound-boxed in the image and the \inference along with its confidence rating. The workers are then asked rate the \observations along three dimensions: (1) acceptability of the \observation (is the \observation reasonable given the image?), (2) appropriateness of bounding boxes (do the bounding boxes appropriately represent the clue?), and (3) interestingness of the \observation (how interesting is the \observation?). The annotation template of the HIT is shown in \fig~\ref{fig:sample_val_hit}.

\subsection{Details on exploration of social biases}
\label{sec:sec_with_bias_investigation_appendix}

The \clues and \inferences we collect from crowdsource workers are abductive, and thus are uncertain. Despite this type of reasoning being an important aspect of human cognition, heuristics and assumptions may reflect false and harmful social biases. As a concrete example: early on in our collection process during a qualifying round, we asked ~70 workers to annotate an image of a bedroom, where action figures were placed on the bed. Many said that the bedroom was likely to belong to a \emph{male} child, citing the action figures as evidence. We again emphasize that our goal is to \emph{study} heuristic reasoning, without endorsing the particular inferences themselves.

\paragraph{Sample analysis.} While curating the corpus, we (the authors) have examined several thousand annotations. To supplement our qualitative experience, in addition, we conducted a close reading of a random sample of 250 inferences. This close reading was focused on references to protected characteristics of people and potentially offensive/NSFW cases.

During both our informal inspection and close reading, we observed similar patterns. Like in other vision and language corpora depicting humans, the most common reference to a protected characteristic was perceived gender, e.g., annotators often assumed depicted people were ``a man" or ``a woman" (and sometimes, age is also assumed, e.g., ``an old man"). Aside from perception standing-in for identity, a majority of \inferences are not specifically/directly about protected characteristics and are SFW (243/250 in our sample). The small number of exceptions included: assumptions about the gender of owners of items similar to the action figure example above (1/250 cases); speculation about the race of an individual based on a sweater logo (1/250); and commenting on bathing suits with respect to gender (1/250). %

Since still frames in VCR are taken from movies, some depict potentially offensive imagery, e.g., movie gore, dated tropes, etc. The images in VCR come with the following disclaimer, which we also endorse (via \href{https://visualcommonsense.com/download/}{visualcommonsense.com}): ``many of the images depict nudity, violence, or miscellaneous problematic things (such as Nazis, because in many movies Nazis are the villains). We left these in though, partially for the purpose of learning (probably negative but still important) commonsense implications about the scenes. Even then, the content covered by movies is still pretty biased and problematic, which definitely manifests in our data (men are more common than women, etc.)."

\paragraph{Statistical analysis.} While the random sample analysis suggests that a vast majority of annotations in our corpus do not reference protected characteristics and are SFW, for an additional check, we passed a random set of 30K samples (10K each from training/val/test) \clues/\inferences through the Perspective API.\footnote{\url{https://www.perspectiveapi.com/}; November 2021 version.} While the API itself is imperfect and itself has biases \cite{hosseini2017deceiving,mitchell2019model,sap2019risk}, it nonetheless can provide some additional information on potentially harmful content in our corpus. We examined the top 50 \clue/\inference pairs across each split marked as most likely to be toxic. Most of these annotations were false positives, e.g., ``a dirty spoon" was marked as potentially toxic likely because of the word ``dirty." But, this analysis did highlight a very small amount of lewd/NSFW/offensive content. Out of the 30K cases filtered through the perspective API, we discovered 6 cases of weight stigmatization, 2 (arguably) lewd observation, 1 dark comment about a cigarette leading to an early death for a person, 1 (arguable) case of insensitivity to mental illness, 6 cases of sexualized content, and 1 (arguable) case where someone was highlighted for wearing non-traditionally-gendered clothing.

\section{Additional Modeling Details}

\label{sec:sec_with_modeling_details}

After some light hyperparameter tuning on the validation set, the best learning rate for fine-tuning our CLIP models was found to be .00001 with AdamW \cite{loshchilov2017decoupled,kingma2014adam}. We use a linear learning rate warmup over 500 steps for \texttt{RN50x16} and \texttt{ViT-B/16}, and 1000 for \texttt{RN50x64}. Our biggest model, \texttt{RN50x64}, takes about 24 hours to converge when trained on 8 Nvidia RTX6000 cards. For data augmentation during training,
we use \texttt{pytorch}'s
\texttt{RandomCrop},
\texttt{RandomHorizontalFlip},
\texttt{RandomGrayscale},
and \texttt{ColorJitter}. For our widescreen CLIP variants, data augmentations are executed on each half of the image independently. We compute visual/textual embeddings via a forward pass of the respective branches of CLIP --- for our widescreen model, we simply average the resultant embeddings for each side of the image. To compute similarity score, we use cosine similarity, and then scale the resulting similarities using a logit scaling factor, following \cite{radford2021learning}. Training is checkpointed every 300 gradient steps, and the checkpoint with best validation $P@1$ \retrieval performance is selected.

\paragraph{Ablation details.} For all ablations, we use the \texttt{ViT-B/16} version of CLIP for training speed: this version is more than twice as fast as our smallest ResNet, and enabled us to try more ablation configurations.

\paragraph{A cleaner training corpus.} Evaluations are reported over version 1.1 of the Sherlock validation/test sets. However, our models are trained on version 1.0, which contains 3\% more data; early experiments indicate that the removed data doesn't significantly impact model performance. This data was removed because we discovered a small number of annotators were misusing the original collection interface, and thus, we removed their annotations. We encourage follow-up work to use version 1.1, but include version 1.0 for the sake of replicability.

\paragraph{T5 model details.} We train T5-Large to map from \clues to \inferences using the Huggingface transformers library \cite{wolf-etal-2020-transformers}; we parallelize using the Huggingface \texttt{accelerate} package. We use Adafactor \cite{shazeer2018adafactor} with learning rate .001 and batch size 32, train for 5 epochs, and select the checkpoint with the best validation loss. 

\subsection{Results on \Clues instead of \Inferences}
\label{sec:sec_with_clue_results}

\begin{table}[t]
\centering
\resizebox{.95\linewidth}{!}{
\begin{tabular}{lccc}
  & \multicolumn{2}{c}{\emph{\Retrieval}} & \emph{\Localization} \\
  \cmidrule(lr){2-3}\cmidrule(lr){4-4}
  & im $\rightarrow$ txt ($\downarrow$) & $P@1_{im \rightarrow txt}$ ($\uparrow$) & GT-Box/Auto-Box ($\uparrow$)  \\
  \midrule
  \texttt{RN50x64}-\inference & 12.8 & 43.4 & 92.5/41.4 \\
  \texttt{RN50x64}-\clue & 6.2 & 54.3 & 94.7/53.3 \\
  \texttt{RN50x64}-multitask & \textbf{5.4} & \textbf{57.5} & \textbf{95.3}/\textbf{54.3} \\
  \bottomrule
\end{tabular}
}
\caption{\Retrieval and \localization results when \clues are used at evaluation time instead of \inferences. This task is more akin to referring expression \retrieval/\localization rather than abductive commonsense reasoning. While \clue \retrieval/\localization setups are easier overall (i.e., referring expressions are easier both models to reason about) the model trained for abductive reasoning, \texttt{RN50x64}-\inference, performs worse than the model trained on referring expressions \texttt{RN50x64}-\clue.}
\label{tab:clues_instead_results}
\end{table}

Whereas \inferences capture abductive inferences, \clues are more akin to referring expressions. While \inferences are our main focus at evaluation time, \sherlock also contains an equal number of \clues, which act as literal descriptions of image regions: \sherlock thus provides a new dataset of 363K localized referring expressions grounded in the image regions of VisualGenome and VCR. As a pointer towards future work, we additionally report results for the \retrieval and \localization setups, but instead of using a version testing on \inference texts, we test on \clues. We do not report over our human-judged \comparison sets, because or raters only observed \inferences in that case. Table~\ref{tab:clues_instead_results} includes prediction results of two models in this setting: both are \texttt{RN50x64} models trained with widescreen processing and with clues highlighted in pixel space, but one is trained on \inferences, and one is trained on \clues.

\section{Batch Size Ablation}

\label{sec:sec_with_batch_size_ablation}

We hypothesize the nature of the hard negatives the models encounter during training is related to their performance.
Because UNITER and LXMERT are bidirectional, they are quadratically more memory intensive vs. CLIP: as a result, for those models, we were only able to train with
18 
negative examples per positive (c.f.  CLIP \texttt{ViT-B/16}, which uses 511 negatives). %
To check that batch size/number of negatives wasn't the only reason CLIP outperformed UNITER, we conducted an experiment varying \texttt{ViT-B/16}'s batch size from 4 to 512; the results are given in \fig~\ref{fig:batch_size}.
Batch size doesn't explain all performance differences: with a batch size of only 4, our weakest CLIP-based model still localizes better than UNITER, and, at batch size 8, it surpasses UNITER's \retrieval performance. %

\section{\Clues and \inferences vs. literal captions}

\begin{figure}[!ht]
    \begin{minipage}{.45\linewidth}
        \centering
        \includegraphics[width=0.9\linewidth]{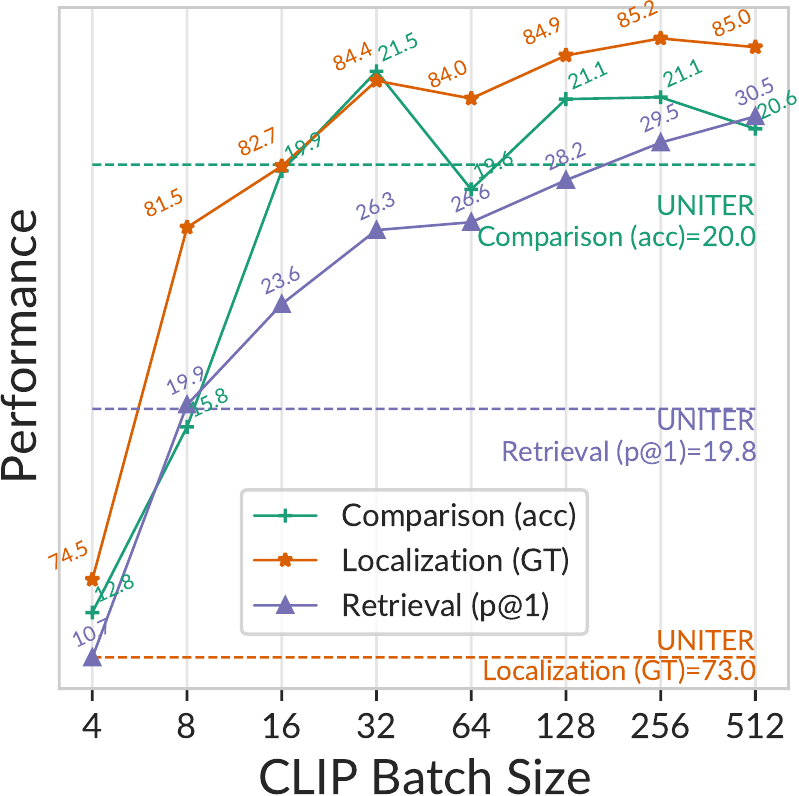}
        \captionof{figure}{The effect of batch size on performance of \texttt{ViT/B-16}. UNITER batch size is 256. Performance on all tasks increases with increasing batch size, but appears to saturate, particularly for \comparison.
        }
        \label{fig:batch_size}
     \end{minipage}
     \begin{minipage}{.49\linewidth}
        \centering
        \includegraphics[width=\linewidth]{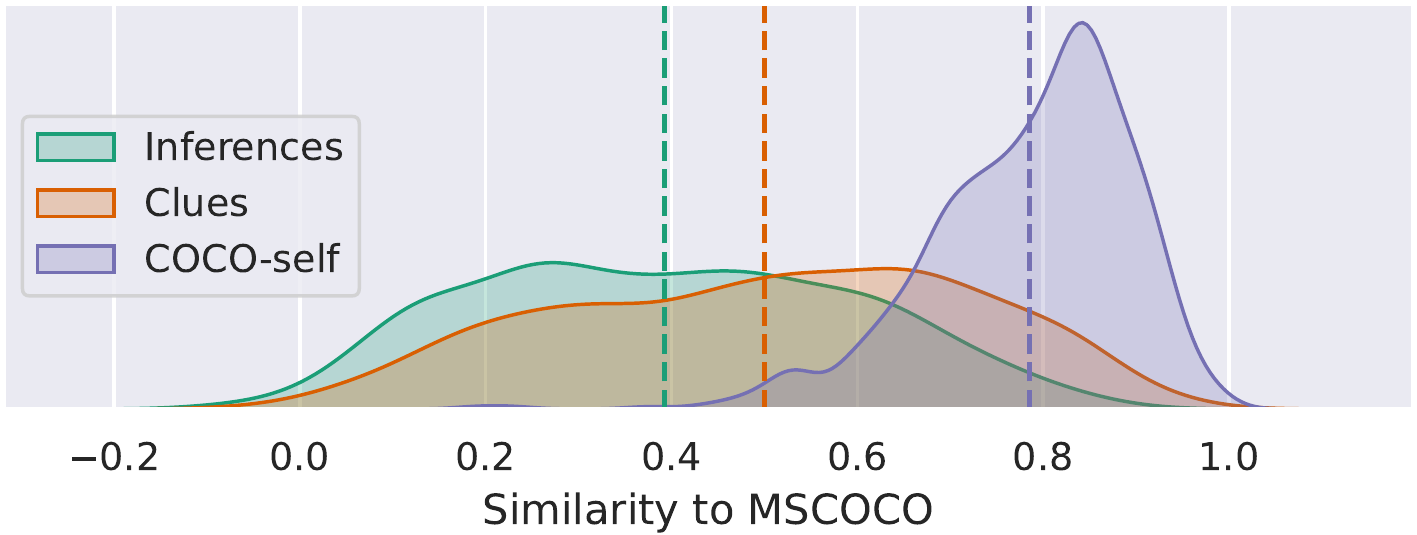}
        \captionof{figure}{The SentenceBERT \cite{reimers-2019-sentence-bert} cosine similarity between \clues/\inferences and MSCOCO captions; MSCOCO caption self-similarity included for reference. On average, \clues are closer to MSCOCO captions than \inferences.}
        \label{fig:overlap_from_mscoco}
    \end{minipage}
\end{figure}

\label{sec:sec_with_overlap_mscoco}

We ran additional analyses to explore the textual similarity between
\sherlock's \clues and \inferences vs. literal image descriptions. For 2K images, we computed text overlap using S-BERT cosine similarity \cite{reimers-2019-sentence-bert} between MS~COCO captions and \sherlock clues/inferences. The result is in \fig~\ref{fig:overlap_from_mscoco}. As a baseline we include COCO self-similarity with held-out captions. Clues are more similar to COCO captions than inferences, presumably because they make reference to the same types of literal objects/actions that are described in literal captions.
\section{\Comparison Human Evaluation Set Details}

\label{sec:sec_w_human_evaluation_details}

We aim to sample a diverse and plausible set of candidate \inferences for images to form our \comparison set.
Our process is a heuristic effort designed to elicit ``interesting" annotations from human raters. Even if the process isn't perfect for generating interesting candidates, because we solicit human ratings we show \inferences to annotators and ask them to rate their plausibility, the resulting set will still be a valid representation of human judgment. We start by assuming all \inferences could be sampled for a given image+region, and proceed to filter according to several heuristics.

First, we use a performant \texttt{RN50x16} checkpoint as a means of judging plausibility of inferences. This checkpoint achieves 18.5/20.6/31.5 im2txt/txt2im/P@1 respectively on \retrieval on v1.0 of the \sherlock corpus; this is comparable to the \texttt{RN50x16} checkpoint we report performance on in our main results section. We use this checkpoint to score all validation/test (image+region, \inference) possibilities.

\paragraph{Global filters.} We assume that if the model is already retrieving its ground truth \inference which high accuracy, the instance is probably not as interesting: for each image, we disqualify all \inferences that receive a lower plausibility estimate from our \texttt{RN50x16} checkpoint vs. the ground truth \inference (this also discards the ground-truth inference). This step ensures that the negative \inferences we sample are more plausible than the ground truth \inference according to the model. Next, we reduce repetitiveness of our \inference texts using two methods. First, we perform the same semantic de-duplication via hierarchical clustering as described in \S~\ref{sec:dataset-collection}: clustering is computed on SentenceBERT \cite{reimers-2019-sentence-bert} representations of inferences (\texttt{all-MiniLM-L6-v2}). We compute roughly 18K clusters (corresponding to 80\% of the dataset size) and sample a single inference from each cluster: this results in 20\% of the corpus being removed from consideration, but maintains diversity, because each of the 18K clusters is represented. Second, we perform a hard-deduplication by only allowing three verbatim copies of each \inference to be sampled.

\paragraph{Local filters.} After these global filters, we begin the iterative sampling process for each image+region. If, after all filtering, a given image+region has fewer than 20 candidates to select from, we do not consider it further. Then, in a greedy fashion, we build-up the candidate set by selecting the remaining \inference with i) the highest model plausibility ii) that is maximally dissimilar to the already sampled \inferences for this image according to the SentenceBERT representations. Both of these objectives are cosine similarities in vector spaces (one between image and text, and one between text and text). We assign weights so that the image-text similarity (corresponding to \texttt{RN50x16} plausibility) is 5x more important than the text-text dissimilarity (corresponding to SentenceBERT diversity). After iteratively constructing a diverse and plausible set of 10 \inferences for a given image under this process, we globally disqualify the sampled \inferences such that no \inference is sampled more than once for each image (unless it is a verbatim duplicate, in which case, it may be sampled up to 3 times).

Finally, for all of the images we are able to sample a set of 10 inferences for, we sort by how promising they are collectively according to a weighted sum of: the (globally ranked) average length of the sampled inferences, the (globally ranked) diversity of the set of 10 (measured by mean all-pairs SentenceBERT cosine sim: lower=more diverse), and 5x the (globally ranked) average plausibility according to \texttt{RN50x16}. We collect 2 human judgments for each of the 10 \inferences for the top 500 images from the val/test sets (1K total) according to this heuristic ranking. The total is 20K human judgments, which formed v1 of the \sherlock \comparison corpus. v1.1 has 19K judgments.

\begin{figure*}

\begin{center}
\includegraphics[width=0.9\linewidth]{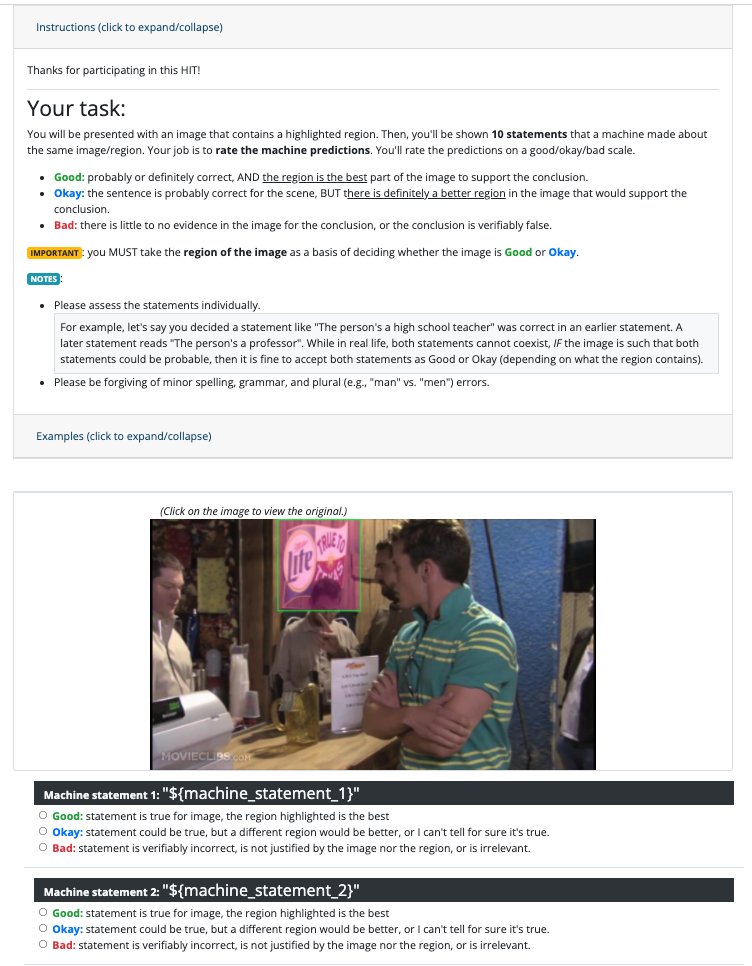}
\end{center}

\caption{
Instructions and template setup for \sherlock model evaluation HIT.
}

\label{fig:sample_eval_hit}
\end{figure*}

\paragraph{Crowdowrking details.}
For the \comparison task, we designed an additional HIT to collect human feedback on the retrieved \inferences. In the HIT, workers were presented with the images with the appropriate clue region highlighted. Then they were provided with the inferences and were asked to rate them on a likert scale of 1-3, with 1 as ``irrelevant'' or ``verifiably incorrect'', 2 as ``statement is probably true but there is a better highlighted region to support it'', and  3 as ``statement is probably true and the highlighted region supports it''. A sample of evaluation HIT is shown in \fig~\ref{fig:sample_eval_hit}. Human agreement on this setup is reported as accuracy \S\ref{sec:sec_with_results}.

\section{Datasheet for \sherlock}

\label{sec:sec_with_datasheet}

In this section, we present a Datasheet \cite{gebru2018datasheets,bender2018data} for \sherlock.

\begin{enumerate}[leftmargin=*,topsep=0pt,itemsep=-1ex,partopsep=1ex,parsep=1ex]

\item Motivation For Datasheet Creation
\begin{itemize}[leftmargin=*,topsep=0pt,itemsep=-1ex,partopsep=1ex,parsep=1ex]
  \item \textbf{Why was the dataset created?} \sherlock was created to support the study of visual abductive reasoning. Broadly speaking, in comparison to corpora which focus on concrete, objective facets depicted within visual scenes (e.g., the presence/absence of objects), we collected \sherlock with the goal of better understanding the types of abductive inferences that people make about images. All abductive inferences carry uncertainty. We aim to study the inferences we collect, but do not endorse their objectivity, and do not advocate for use cases that risk perpetuating them.
  
  \item \textbf{Has the dataset been used already?} The annotations we collect are novel, but the images are sourced from two widely-used, existing datasets: Visual Genome \cite{Krishna2016VisualGC} and VCR \cite{zellers2019recognition}.
  
\item \textbf{What (other) tasks could the dataset be used for?} Aside from our retrieval/localization setups, \sherlock could be useful as a pretraining corpus for models that aim to capture information about what people might assume about an image, rather than what is literally depicted in that image. One potentially promising case: if a malicious actor were posting emotionally manipulative content online, it might be helpful to study the types of assumptions people might make about their posts, rather than the literal contents of the post itself. 

\item \textbf{Who funded dataset creation?}
This work was funded by DARPA MCS program through NIWC Pacific (N66001-19-2-4031), the DARPA SemaFor program, and the Allen Institute for AI.
\end{itemize}

\item Data composition
\begin{itemize}[leftmargin=*,topsep=0pt,itemsep=-1ex,partopsep=1ex,parsep=1ex]
\item \textbf{What are the instances?} We refer to the instances as \clues/\inferences, which are authored by crowdworkers. As detailed in the main text of the paper, a \clue is a bounding box coupled with a free-text description of the literal contents of that bounding box. An \inference is an abductive conclusion that the crowdworker thinks could be true about the \clue.

\item \textbf{How many instances are there?} There are 363K commonsense inferences grounded in 81K Visual Genome images and 22K VCR images.

\item \textbf{What data does each instance consist of?} Each instance contains 3 things: a \clue, a short English literal description of a portion of the image, an \inference, a short English description of an \inference associated with the clue that aims to be not immediately obvious from the image content, and a bounding box specified with the region of interest.

\item \textbf{Is there a label or target associated with each instance?} We discuss in the paper several tasks, which involve predicting \inferences, bounding boxes, etc.

\item \textbf{Is any information missing from individual instances?} Not systematically --- in rare circumstances, we had to discard some instances because of malformed crowdworking inputs.

\item \textbf{Are relationships between individual instances made explicit?} Yes --- the annotations for a given image are all made by the same annotator and are aggregated based on that.

\item \textbf{Does the dataset contain all possible instances or is it a sample?} This is a natural language sample of abductive \inferences; it would probably be impossible to enumerate all of them.

\item \textbf{Are there recommended data splits?} Yes, they are provided.

\item \textbf{Are there any errors, sources of noise, or redundancies in the dataset? If so, please provide a description.} Yes: some annotations are repeated by crowdworkers. When we collected the corpus of Likert judgments for evaluation, we performed both soft and hard deduplication steps, ensuring that the text people were evaluating wasn't overly repetitive.

\item \textbf{Is the dataset self-contained, or does it link to or otherwise rely on external resources (e.g., websites, tweets, other datasets)?} It links to the images provided by Visual Genome and VCR. If images were removed from those corpora, our annotations wouldn't be grounded.

\end{itemize}
\item Collection Process
\begin{itemize}[leftmargin=*,topsep=0pt,itemsep=-1ex,partopsep=1ex,parsep=1ex]

\item \textbf{What mechanisms or procedures were used to collect the data?} We collected data using Amazon Mechanical Turk.

\item \textbf{How was the data associated with each instance acquired? Was the data directly observable (e.g., raw text, movie ratings), reported by subjects (e.g., survey responses), or indirectly inferred or derived from other data?} Paid crowdworkers provided the annotations.

\item \textbf{If the dataset is a sample from a larger set, what was the sampling strategy (e.g., deterministic, probabilistic with specific sampling probabilities)?} We downsample common image types via a semantic deduplication step. Specifically, some of our crowdworkers were rightfully pointing out that it's difficult to say interesting things about endless pictures of zebra; these types of images are common in visual genome. So, we performed hierarchical clustering on the images from that corpus, and then sampled 1 image from each of 80K clusters. The result is a downsampling of images with similar feature representations. We stopped receiving comments about zebras after this deduplication step.

\item \textbf{Who was involved in the data collection process (e.g., students, crowdworkers, contractors) and how were they compensated (e.g., how much were crowdworkers paid)?} Crowdworkers constructed the corpus via a mechanical turk HIT we designed. We our target was to pay \$15/hour. A post-hoc analysis revealed that crowdworkers were paid a median \$12/hr and a mean of \$16-20/hour, depending on the round.

\item \textbf{Over what timeframe was the data collected? Does this timeframe match the creation timeframe of the data associated with the instances (e.g., recent crawl of old news articles)? If not, please describe the timeframe in which the data associated with the instances was created.} The main data was collected in February 2021.

\end{itemize}
\item {Data Preprocessing}
\begin{itemize}[leftmargin=*,topsep=0pt,itemsep=-1ex,partopsep=1ex,parsep=1ex]
\item \textbf{Was any preprocessing/cleaning/labeling of the data done (e.g., discretization or bucketing, tokenization, part-of-speech tagging, SIFT feature extraction, removal of instances, processing of missing values)?} Yes, significant preprocessing was conducted. The details are in 

\item \textbf{Was the ``raw" data saved in addition to the preprocessed, cleaned, labeled data (e.g., to support unanticipated future uses)? If so, please provide a link or other access point to the `raw' data.} The concept of ``raw" data is difficult to specify in our case. We detail the data we release in the main body of the paper.

\item \textbf{Is the software used to preprocess/clean/label the instances available? If so, please provide a link or other access point.} We plan to release some software related to modeling, and also have provided some appendices that detail the crowdworking labelling efforts.

\item \textbf{Does this dataset collection/processing procedure achieve the motivation for creating the dataset stated in the first section of this datasheet? If not, what are the limitations?} We think so. It's difficult to fully specify the abductive reasoning process of humans. But we think our work goes a step beyond existing corpora.

\end{itemize}
\item Dataset Distribution
\begin{itemize}[leftmargin=*,topsep=0pt,itemsep=-1ex,partopsep=1ex,parsep=1ex]
\item \textbf{How will the dataset be distributed?}

The dataset is available at \url{http://visualabduction.com/}.

\item \textbf{When will the dataset be released/first distributed? What license (if any) is it distributed under?}

The dataset is released under CC-BY 4.0 and the code is released under Apache 2.0.

\item \textbf{Are there any copyrights on the data?}

The copyright for the new annotations is held by AI2 with all rights reserved.

\item \textbf{Are there any fees or access restrictions?}

No --- our annotations are freely available.

\end{itemize}
\item Dataset Maintenance
\begin{itemize}
\item \textbf{Who is supporting/hosting/maintaining the dataset?}

The dataset is hosted and maintained by AI2.

\item \textbf{Will the dataset be updated? If so, how often and by whom?}

We do not currently have plans to update the dataset regularly.

\item \textbf{Is there a repository to link to any/all papers/systems that use this dataset?}

No, but if future work finds this work helpful, we hope they will consider citing this work.

\item \textbf{If others want to extend/augment/build on this dataset, is there a mechanism for them to do so?}

People are free to remix, use, extend, build, critique, and filter the corpus: we would be excited to hear more about use cases either via our github repo, or via personal correspondence.

\end{itemize}
\item Legal and Ethical Considerations
\begin{itemize}

\item \textbf{Were any ethical review processes conducted (e.g., by an institutional review board)?}

Crowdworking studies involving no personal disclosures of standard computer vision corpora are not required by our IRB to be reviewed by them. While we are not lawyers, the opinion is based on United States federal regulation 45 CFR 46, under which this study qualifies and as exempt and does not require IRB review. 

\begin{enumerate}
\item We do not collect personal information. Information gathered is strictly limited to general surveys probing at general world knowledge.\vspace{1.1mm}

\item We take precaution to anonymize Mechanical WorkerIDs in a manner that the identity of the human subjects cannot be readily ascertained (directly or indirectly).\vspace{1.1mm}

\item We do not record or include any interpersonal communication or contact between investigation and subject.

\end{enumerate}

Specifically:

\begin{itemize}
\item We do not have access to the underlying personal records and will record information in such a manner that the identity of the human subject cannot readily be ascertained. 
\vspace{1mm}
\item Information generated by participants is non-identifying without turning over the personal records attached to these worker IDs.
\vspace{1mm}
\item We do not record or include any interpersonal communication or contact between investigation and subject.
\end{itemize}

\item \textbf{Does the dataset contain data that might be considered confidential?}

Potentially, yes. Most of the content in the corpus that would be considered potentially private/confidential would likely be depicted in the images of Visual Genome (VCR are stills from movies where actors on-screen are presumably aware of their public actions). While we distribute no new images, if an image is removed from Visual Genome (or VCR), it will be removed from our corpus as well.

\item \textbf{Does the dataset contain data that, if viewed directly, might be offensive, insulting, threatening, or might otherwise cause anxiety? If so, please describe why}

As detailed in the main body of the paper, we have searched for toxic content using a mix of close reading of instances and the Perspective API from Google. In doing this, we have identified a small fraction of instances that could be construed as offensive. For example, in a sample of 30K instances, we discovered 6 cases that arguably offensive (stigmatizes depicted people's weight based on visual cues). Additionally, some of the images from VCR, gathered from popular movies, can depict potentially offensive/disturbing content. The screens can be ``R Rated," e.g., some images depict movie violence with zombies, some of the movies have Nazis as villains, and thus, some of the screenshots depict Nazi symbols. We reproduce VCR's content warning about such imagery in \S~\ref{appendix:data-collection}.

\item \textbf{Does the dataset relate to people?}

Yes: the corpus depicts people, and the annotations are frequently abductive inferences that relate to people. As detailed in the main body of the paper, 36\% of \inferences (or more) are grounded on people; and, many inferences that are not directly grounded on people may relate to them. Moreover, given that we aim to study abduction, which is an intrinsically subjective process, the annotations themselves are, at least in part, reflections of the annotators themselves.

\item \textbf{Does the dataset identify any subpopulations (e.g., by age, gender)?}

We don't explicitly disallow identification by gender or age, e.g., in the \clues/\inferences, people often will use gendered pronouns or aged language in reference to people who are depicted (e.g., ``the old man"). Furthermore, while we undertook the sample/statistical toxicity analysis detailed in the main body of the paper, we have not manually verified that all 363K \clue/\inference pairings are free of any reference to a subpopulation. For example, we observed one case wherein an author speculated about the country-of-origin of an individual being Morroco, clued by the observation that they were wearing a fez. Like the other observations in our corpus, it's not necessarily the case that this is an objectively true inference, even if the fez is a hat that is worn in Morroco.

\item \textbf{Is it possible to identify individuals (i.e., one or more natural persons), either directly or indirectly (i.e., in combination with other data) from the dataset?}

The data collection process specifically instructs workers to avoid identifying any individual in particular (e.g., actors in movie scenes). Instead, they are specifically instructed to use general identifiers to describe people (e.g. ``student'', ``old man'', ``engineer''). 
In our experience with working with the corpus, we haven't encountered any instances where our annotators specifically identified anyone, e.g., by name. The images contained in VCR and Visual Genome that we source from do contain uncensored images of faces. But, if images are removed from those corpora, they will be removed from Sherlock as well, as we do not plan to re-host the images ourselves.

\end{itemize}
\end{enumerate}

\end{document}